\documentclass[preprint]{elsarticle}

\usepackage{lineno,hyperref}
\usepackage[normalem]{ulem}
\usepackage{seqsplit}
\RequirePackage{fix-cm}
\usepackage{subcaption}
\usepackage[normalem]{ulem}
\usepackage{amssymb}
\usepackage{amsthm}

\modulolinenumbers[5]

\journal{Journal of Applied Soft Computing}

\usepackage{mathtools}
\usepackage{commath}
\usepackage{algorithm}
\usepackage{algpseudocode}

\usepackage{multirow}
\usepackage{amsmath}
\usepackage{amssymb}
\usepackage{graphicx}
\usepackage{xcolor}
\graphicspath{./plot}

\DeclareMathOperator{\atantwo}{atan2}

\algnewcommand\algorithmicforeach{\textbf{for each}}
\algdef{S}[FOR]{ForEach}[1]{\algorithmicforeach\ #1\ \algorithmicdo}

\newcommand{\NEO}{\textcolor{black}{neuroevolution}}

\newcommand{\NEOT}{\textcolor{black}{neurotrajectory prediction}}

\begin{document}
	
	\begin{frontmatter}
		
		\title{Evolutionary Multi-objective Optimisation \\in Neurotrajectory Prediction}

		\author{Edgar Galv\'{a}n \corref{firstauthor} \corref{mycorrespondingauthor} (ORCID: 0000-0001-8474-5234)}
		\address{Naturally Inspired Computation Research Group, Department of Computer Science, Hamilton Institute, National University of Ireland Maynooth, Lero, Ireland}
		\ead{edgar.galvan@mu.ie}

		\author{Fergal Stapleton \corref{firstauthor} (ORCID: 0000-0002-5347-1573)}
		\address{Naturally Inspired Computation Research Group, Department of Computer Science, Hamilton Institute, National University of Ireland Maynooth, Ireland}
		\ead{fergal.stapleton.2020@mumail.ie}
		
		\cortext[firstauthor]{Joint first authors.}
		\cortext[mycorrespondingauthor]{Corresponding author.}
		
		\begin{abstract}
			
			Machine learning has rapidly evolved during the last decade, achieving expert human performance on notoriously challenging problems such as image classification. This success is partly due to the re-emergence of bio-inspired modern artificial neural networks (ANNs) along with the availability of computation power, vast labelled data and ingenious human-based expert knowledge as well as optimisation approaches that can find the correct configuration (and weights) for these networks.  Neuroevolution is a term used for the latter when employing evolutionary algorithms. Most of the works in neuroevolution have focused their attention in a single type of ANNs, named Convolutional Neural Networks (CNNs). Moreover, most of these works have used a single optimisation approach. This work makes a progressive step forward in neuroevolution for vehicle trajectory prediction, referred to as \NEOT, where multiple objectives must be considered. To this end, rich ANNs composed of CNNs and Long-short Term Memory Network are adopted. Two well-known and robust Evolutionary Multi-objective Optimisation (EMO) algorithms, named Non-dominated Sorting Genetic Algorithm-II (NSGA-II) and Multi-Objective Evolutionary Algorithm with Decomposition (MOEA/D) are also adopted. The completely different underlying mechanism of each of these algorithms sheds light on the implications of using one over the other EMO approach in \NEOT. In particular, the importance of considering objective scaling is highlighted, finding that MOEA/D can be more adept at focusing on specific objectives whereas, NSGA-II tends to be more invariant to objective scaling. Additionally, certain objectives are shown to be either beneficial or detrimental to finding valid models, for instance, inclusion of a distance feedback objective was considerably detrimental to finding valid models, while a lateral velocity objective was more beneficial.
			

		\end{abstract}
		
		
	\end{frontmatter}
	
	
	\section{Introduction}
	\label{sec:introduction}
	
	Modern bio-inspired artificial neural networks (ANNs)~\citep{Goodfellow-et-al-2016,DBLP:journals/nature/LeCunBH15}, a subset of machine learning algorithms are inspired by deep hierarchical structures of human perception as well as production systems~\cite{galvan2020neuroevolution}. These have achieved human level performance in multiple areas from image classification~\citep{DBLP:conf/cvpr/SzegedyLJSRAEVR15}, face recognition~\cite{DBLP:conf/gecco/DengSG22,Yanan2023} to playing complex games of intractable search and decision spaces~\citep{Silver_2016}. Evolutionary Algorithms~\citep{Back:1996:EAT:229867,EibenBook2003}, also known as Evolutionary Computation Systems, are nature-inspired stochastic optimisation techniques that mimic basic principles of life. These automatic algorithms have been with us for several decades and are highly popular given that they have proven competitive in the face of challenging problems’ features such as discontinuities, multiple local optima, non-linear interactions between variables, deceptiveness, among other characteristics~\citep{Eiben:2015:nature}. They have also proven to yield competitive results in multiple real-world problems against other Artificial Intelligent methods as well as results achieved by human experts~\citep{koza_2003,Koza:2010:HRP:1831229.1831232}.
	
	The intertwined use of these two revolutionary bio-inspired algorithms was common in the early studies of training and learning in ANNs by using evolutionary algorithms (EAs). However, both communities became fragmented more than three decades ago mainly as a result of the ANNs community finding an elegant solution to the training of ANNs~\citep{Bottou2012,HECHTNIELSEN199265,726791}. These areas have witnessed a re-emerge of working together thanks to the use of EAs in the automatic configuration of these networks and, to a lesser degree, in their training, commonly referred to as neuroevolution in ANNs. This is described in detail in the first author's IEEE Transactions on AI article~\cite{galvan2020neuroevolution}, reviewing over 170 recent papers published in the last seven years, discussing and evaluating state-of-the-art neuroevolution approaches.
	
	
	Surprisingly, most of the works in \NEO\  have focused their attention on single optimisation. Moreover, the vast majority of these ingenious studies have concentrated their attention on ANNs: Convolutional Neural Networks (CNNs)~\cite{10.1145/3065386}. In sharp contrast to this tendency, this work focuses its attention on the challenging task of autonomous driving through trajectory prediction. To this end, a rich ANN, composed of a CNN and a Long-short Term Memory (LSTM) network is used~\cite{iet:/content/conferences/10.1049/cp_19991218}. These two are well-established networks within the ANN community. In this work, neuroevolution is employed to automatically find the right configuration of the hyperparameters’ values of this network. This process is referred to as \NEOT. Multiple objectives are considered in this study, some of which are in clear conflict with each other.  To this end,  two well-known and robust Evolutionary Multi-objective Optimisation (EMO)~\cite{Deb:2001:MOU:559152} approaches, namely Multi-Objective Evolutionary Algorithm with Decomposition (MOEA/D)~\cite{4358754} and Non-dominated Sorting Genetic Algorithm-II (NSGA-II)~\cite{996017} are used. These two popular algorithms are used given their distinct underlying mechanism. One is based on decomposition considering a weighted mechanism to carry out evolutionary search (MOEA/D). The other is based on a Pareto-based highly popular MO approach (NSGA-II).

	
	
	The goal of this paper is to investigate how the key underlying mechanisms adopted in multiple EMO approaches such as decomposition and Pareto dominance-based techniques behave in the challenging task of neurotrajectory. To this end, we use two well-known algorithms that adopt each of these underlying mechanisms, named MOEA/D and NSGA-II. The underlying mechanisms of these two approaches are representative for a number of other EMOs, for instance, MOEA/D based variants using weighted decomposition~\cite{8998284} or dominance-based approaches like SPEA2~\cite{Zitzler2001SPEA2IT}. Furthermore, they are two of the most popular EMO frameworks and have been adopted for many different problem domains~\cite{coello2004applications}. An in-depth comparison of these two EMO frameworks is performed and recommendations are made for when each approach is best suited for the \NEOT\ task. It is important to note, however, that the objective is not to make a blanket statement on which EMO framework is better but rather to provide recommendations to future researchers based on their needs and the specific algorithmic qualities of each EMO framework. Understanding these qualities offers researchers the oportunity for selecting EMOs based on their specific needs, for example, in this work the issue of objective scaling is highlighted, a subject which is of vital importance for researchers working on practical real-word problems, where normalisation may not be guaranteed and which is handled differently by each framework. Because these qualities are inherent to each type of EMO framework, the observations made can not only be generalized for other frameworks with the same underlying mechanism, but also understanding these qualities can be useful for problem-domains outside of the field of trajectory prediction.
	

Therefore, while there have been inspiring neuroevolution studies focusing primarily on single optimisation and CNNs as discussed in~\cite{galvan2020neuroevolution,GalvanSig2021}, the contributions of this work are both novel and timely.

\begin{enumerate}
    \item Firstly, we propose a comprehensive artificial neural network (ANN) architecture comprising a combination of a Convolutional Neural Network (CNN) and a Long Short-Term Memory (LSTM) network. This ANN is designed to extract essential feature information from the sequence of trajectory images using the CNN and then use the LSTM network for accurate trajectory prediction.

\item In addition, we identify and define five distinct objectives for this neurotrajectory task. Three of these objectives pertain directly to the trajectory prediction problem itself, namely distance feedback, lateral velocity, and longitudinal velocity. The remaining two objectives are defined during task execution and include the root mean squared error and a loss function. To effectively handle the potential conflicts between these objectives, we employ the widely recognized Spearman rank-order coefficient as a metric to quantify the degree of conflict. This analysis helps us streamline and reduce the overall number of experiments required, as it helps to reduce testing every possible combination of the aforementioned objectives.

\item To ensure robust and comprehensive experimentation, we conduct a series of 156 independent runs, which includes carrying out ablation studies of Evolutionary Multi-Objective Optimization (EMO) in neuroevolution. This goes beyond the norm in the ANN community, where a single run is often the standard due to computational limitations. By employing a considerable larger number of runs, our statistical methodology enhances the credibility of the analysis and conclusions drawn from this work. Overall, our approach introduces robust experimentation techniques to the field of neuroevolution.

\item It is important to highlight that trajectory prediction in neuroevolution is a challenging area that has received relatively little attention. Furthermore, there is a limited understanding of how to effectively incorporate EMO frameworks in this domain. Therefore, our study addresses this gap by using two distinct EMO approaches, MOEA/D (decomposition-based) and NSGA-II (Pareto dominance-based), each employing different underlying mechanisms. By shedding light on the impact of these EMO approaches in the demanding task of neurotrajectory prediction, we contribute to advancing the understanding and application of EMO techniques in this specific area.

 \end{enumerate}

	The remaining sections of this paper are organised as follows. Section~\ref{sec:related} discusses the related work and how this work makes a leap forward in the understanding of EMO in neuroevolution, specifically in \NEOT. 
	Section~\ref{sec:background}
	discusses Convolutional Neural Networks (CNN), Long-short Term Memory Networks (LSTM), Pareto dominance and a brief description of NSGA-II and MOEA/D. Section~\ref{sec:proposed} outlines the proposed approach concerning trajectory prediction, objectives used and network topology. Section~\ref{sec:experimental} explains the experimental setup. Section~\ref{sec:results} shows results and analysis. Finally, Section ~\ref{sec:conclusions} draws some conclusions.

	
	\section{Related Work on EMO in Neuroevolution}
	\label{sec:related}
	

	Evolutionary Multi-objective Optimisation~\cite{Deb:2001:MOU:559152,1597059,CoelloCoello1999}, has seldom been used in the automatic configuration of DNNs and/or in the optimisation of  their hyperparameters. Works on the latter include the approach proposed by Kim et al.~\cite{Kim2017NEMON}, where the authors used two conflicting objectives, speed and accuracy, to be optimised via an EMO process using the NSGA-II. They reported interesting results for three classification tasks, which included the standard image classification datasets MNIST~\cite{deng2012mnist} and CIFAR-10~\cite{krizhevsky2009learning} as well as the drowsy behaviour recognition dataset. Motivated by the work of Kim et al.~\cite{Kim2017NEMON}, Lu et al.~\cite{10.1145/3321707.3321729} used the same objectives, again using NSGA-II, entitling their approach NSGA--Net. Furthermore, in order to measure speed, Lu et al.~\cite{10.1145/3321707.3321729} empirically tested a number of computational complexity metrics, such as the number of active nodes, number of active connections between the nodes and floating-point operations (FLOPs), to mention a few. The authors reported that FLOPs metric was more accurate and was used as a second objective for optimisation. A major contribution of the authors' work was the incorporation of a bit-string encoding for their genetic algorithm to use, allowing for robust and widely used genetic operators normally adopted in GAs, such as bit-flip mutation and homogeneous crossover. Furthermore, a Bayesian learning scheme is used to search promising solutions from an archive, where new offspring are generated based on Bayesian networks. The same authors extended this work~\cite{lu2020multiobjective} from looking at the CIFAR-10 dataset to include more complex datasets like fashion MNIST~\cite{xiao2017fashion}, SVHN~\cite{netzer2011reading} and CIFAR-100~\cite{krizhevsky2009learning}, to name a few. This work makes use of proxy models to downscale more complex architectures and improve efficiency.

	Elsken et al. \cite{elsken2019openreview} proposed a fast multi-objective framework based on Lamarckian inheritance. Their approach has two main facets, firstly they rely on operators that preserve networks functionality using approximate network morphisms, which removes the need to re-train expensive network models from scratch. Secondly, they propose an algorithm: Lamarckian Evolutionary algorithm for Multi-Objective Neural Architecture DEsign (LEMONADE). This algorithm focuses on cheap to evaluate objectives for a subset of architectures assigning higher probabilities for these networks to be fully evaluated. They demonstrate that LEMONADE is capable of handling complex topologies with skip connections and multiple branches.
	
	The trade-off between network size and accuracy is paramount when considering embedded systems where response time, energy consumption and efficiency are critical. Loni et. al.~\cite{loni2020deepmaker} proposed the DeepMaker framework which is designed to find suitable DNN architectures for embedded systems, while optimizing with respect to multiple objectives. Their method incorporates NSGA-II for optimizing with respect to the number of parameters and error rate of the networks. Furthermore their approach prunes the design space to help compression, while still maintaining an acceptable level of accuracy. By designing their framework specifically for embedded systems, their novel approach highlights the value of using EMO for DNN architecture search in a practical sense.
	
	
	Lu et. al.~\cite{lu2023neural} developed a bench marking framework called EvoXBench, which was used to conduct an in-depth comparison of six EMO algorithms for Neural Network architecture search, albeit for smaller networks than what are currently considered deep. While ground breaking for in its own right, the paper is notable for this work in that it highlights the importance of objective scaling. The authors note that decomposition methods are particularly sensitive to the scaling of objectives, but note that it is not always possible to normalise them. A survey of normalisation approaches is provide in~\cite{he2021survey}.


	For more traditional, less deep ANNS, the blending of neurovolutionary approaches and multi-objective optimization has resulted in a number of notable works in recent years. Mirret and Chua et al.~\cite{miret2022neuroevolution} applied Neuroevolution-Enhanced Multi-objective optimization (NEMO) with the aim of optimizing task performance, memory compression and compute savings for mixed-precision quantization. Their approach divides the population into structurally distinct sub-populations which, based on the quality of these solutions, are use to construct the Pareto frontier for the multi-objective problem. The EMO approach used in this work is based upon the NSGA-III algorithm~\cite{deb2013evolutionary}.
	
	
	While this work has selected two EMOs based on their distinct properties, for a  more general but comprehensive survey on the state-of-the-art in EMOs, the work of Liu et al.~\cite{liu2020multi} gives an in-depth overview. Recent survey by Xu et al.~\cite{8998284} is also a good reference for recent trends in decomposition-based variants.
	
	Relevant to this work is the study carried out by Grigorescu et. al.~\cite{grigorescu2019ieee}. The authors considered three objectives to optimise relating specifically to to the problem domain of localised trajectory prediction, the longitudinal velocity of the ego vehicle, the lateral velocity of the ego vehicle and a distance feedback objective. These objectives are particularly interesting as most literature so far has considered objective relating to the performance of the DNN itself. Their approach was shown to be competitive with state of the art, outperforming two other approaches: Dynamic Window Approach (DWA)~\cite{fox1997dynamic} and End2End supervised learning~\cite{bojarski2016end}. The work was proposed as a multi-objective optimisation problem, however, little detail was offered on the framework they employed and consequently little analysis is offered on the optimisation process. This is addressed by using robust and well established EMO frameworks. Furthermore, there is no analysis on the validity of models found, something that is of vital importance to the research community. This paper served as a baseline for this work in terms of the performance metrics used as well as the objectives considered. Furthermore, the network design was adapted to allow for more hyperparameters to be evolved during the optimisation process.

	The work outlined in this paper considerably expands upon our previous work~\cite{stapleton2022neuroevolutionary}, which had a more limited analysis of objective performance and no comparison of EMO frameworks. 
	
	
	
	\section{Background}
	\label{sec:background}
	
	\subsection{Deep Learning Architecture: Convolutional Neural Networks (CNNs)}

	CNNs have shown an incredible ability to process data that has been constructed in a grid-like fashion. Typically, the network architecture is represented as a series of layers containing one or more planes, where each plane is composed of an array of units. Information can then be passed from a neighbourhood of units of one layer to the next. This concept was directly inspired by Hubel and Wiesels~\cite{doi:10.1113/jphysiol.1962.sp006837} work relating to the perceptron and animal visual cortex. Furthering this idea, the basic principle of a CNN is to convolve a filter or kernel over a layer to produce feature maps, which extract important feature information from the input. Each kernel is represented by an array of weights. The weights are trained using traditional backpropogation methods such as the gradient descent process proposed by LeCun et al.~\cite{Lecun98gradient-basedlearning}. The corresponding layer for each of these kernels is referred to as a convolution layer. Additionally, through a process of sub-sampling or down-sampling, it is possible to reduce the dimensionality of the data as it progresses through the various layers, helping to reduce the number of features to learn. CNNs were ground-breaking in helping to establish deep learning architectures and have successfully been applied to numerous domains such as face detection, handwriting recognition, image classification, speech recognition, natural language processing and recommender systems~\cite{10.1145/3065386,dos-santos-gatti-2014-deep,10.1145/3219819.3219890}. 
	
	\subsection{Deep Learning Architecture: Long-short Term Memory Network (LSTM)}

	Long-short-term memory (LSTM) networks belong to the broader class of Recurrent Neural Networks (RNNs) and by design attempt to learn the long-term dependencies of underlying data. LSTM networks were originally developed to tackle the problem of vanishing gradients which had been a problem in other RNN networks. They are particularly effective when used on sequential or time series data and have been applied to a wide range of problems~\cite{7508408}. There are several main components of an LSTM: the cell state, the hidden state, the forget gate, the input gate and the output gate. The cell state is responsible for remembering values over arbitrarily long time intervals, and can be considered as the aggregate memory of the LSTM network across all time-steps. The hidden state is responsible for encoding information from just the previous time-step and is in effect responsible for the short term memory aspect of the network. The next three components are recurrent gates that control the flow of information in and out of the cell state. The forget gate which is responsible for telling the cell state which information to forget. The input gate which is responsible for updating the current step with relevant information and the output gate which is responsible for determining the value of the next hidden state. The weights in the forget and input gate are responsible for determining which time steps are important, as such, help retain or remember the long term dependencies of the cell state. Gers and Schmidhuber~\cite{963769} showed that standard RNNs fail to learn when dealing with time lags in excess of as few as five to ten discrete-time steps between relevant input events and target signals. On the other hand, LSTMs are not affected by this problem and are capable of dealing with time lags far in excess of 1000 discrete-time steps. LSTM networks clearly outperform previous RNNs not only on regular language benchmarks but also on context-free languages benchmarks~\cite{963769}. 
	
	\subsection{EMO: Non-dominated Sorting Genetic Algorithm II}

	In multi-objective optimisation the goal is to find the best trade-off between solutions, with the aim of either minimising or maximising multiple objectives that are conflicting. For minimisation, 
	
	\begin{equation}
	min(f_1(x), f_2(x), ..., f_m(x))\ \ \ \ \ s.t. \ \ x \in X,
	\label{eqn:mo}
	\end{equation}
	
	\noindent where each $f(x)$ is an objective function to be minimised and $x$ is a candidate solution from the feasible solution set $X$, where the number of objectives $m$ is greater or equal than two. Conflicting objectives will result in an infeasible region within the objective space. Thus, candidate solutions are sought which best approximate the boundary between the feasible search space and infeasible search space.

	The Pareto dominance relationship can be used to converge towards this boundary. Pareto dominance can be defined as follows:
	
	\textbf{Def 1:} A candidate solution $x_1$ is said to dominate another candidate solution $x_2$, if

	\begin{equation} 
	f_{i}(x_1) \leq f_i(x_2)\ \forall   i \in \{1, 2, ... m\} \end{equation}
	\begin{equation} 
	f_j(x_1) < f_j(x_2)\ | \ \exists j \in \{ 1, 2, ... m\} 
	\end{equation}\\

	The Non-dominated Sorting Genetic Algorithm II (NSGA-II)~\cite{996017} uses the Pareto dominance relationship as its primary mechanism for selecting solutions to be retained into future generations. It does so by sorting non-dominated solutions based on their dominance rank. More formally, the dominance rank $D$ of a potential solution $S_i$ is determined by how many individual solutions dominate $S_i$. In mathematical terms this can be represented as the cardinality of the set $|.|$ such that,
	
	\begin{equation}
	D(S_i) = |\{j|j \in Pop \land S_j \succ S_i \}|
	\label{eqn:dom_rank}
	\end{equation}

	In this sense, each individual may be assigned a rank and individuals of the same rank will belong to the same front $F_D$. For instance, individuals which have a Dominance Rank of $0$ (i.e., are dominated by no other individual) will belong to front $F_0$, individuals which have a Dominance Rank of $1$ will belong to the second front $F_1$ and so forth. A new population is then generated from these fronts, where the new population size is the same size as the original parent population size.
	
	Since the original population size is unlikely to be exactly filled by the last front, a secondary sort is required to select individuals from the last front and is based on the crowding distance operator. The crowding distance is the sum of Manhattan distances between nearest neighbours of an individual under consideration~\cite{minkowski1910geometrie}. This is one of the most popular and still widely used  EMO approach in the specialised literature as demonstrated by recent publications~\cite{DBLP:conf/gecco/GalvanS19,DBLP:journals/asc/GalvanTS22,DBLP:conf/ssci/GalvanS20}.

	\subsection{EMO: Multi-Objective Evolutionary Algorithm with Decomposition}
	
	The Multi-objective Evolutionary Algorithm with Decomposition \sloppy{(MOEA/D)}~\cite{4358754} works by decomposing a multi-objective optimisation problem into a number of scalar optimisation sub-problems, where each of these exploits a different region of the objective space via weights~\cite{4358754}. These scalar optimisation sub-problems are defined by the scalar optimisation function $g$, which under the canonical approach uses a uniform distribution of weight vectors $\lambda^i$, where $i = \{1, 2 ... m\}$ and $\sum{\lambda^i} = 1$, such that
	
	\begin{equation}
	\lambda^i \in \{ 0, \frac{1}{H} , \frac{1}{H} ... \frac{H}{H} \} \ \ for\ i = \{1, 2 ... m\}
	\label{eq:weights}
	\end{equation}
	
	\noindent where $H$ is an integer value determining the distribution of weights. When considering three-objective problem as done in this work, the weights are equidistantly distributed over a hyperplane in the form of a $\{3, H\}$ simplex-lattice, where the population size is defined as $N = \frac{(H+1)(H+2)}{2}$.
	The neighbour solutions are selected based on the Euclidean distance between the weights in objective space.

	A core mechanism of the decomposition approach is the neighbourhood structure, where genetic operations and the selection process occur within neighbourhoods of closely related sub-problems. In this way, the neighbourhood structure controls the exploration and exploitation properties of the algorithm. In this work, the Tchebycheff approach is used, briefly discussed next. \\
	
	
	
	

	\noindent \textbf{Tchebycheff}\\
	
	The Tchebycheff scalar optimisation $g^{tch}$ is determined by Eq.~\ref{eq:tche}
	
	\begin{equation}
	\min(g^{tch}(x|\lambda)) = \underset{1 \leq j \leq m}{\max}  \{\lambda | f_j(x) - z_{j} | \}
	\label{eq:tche}
	\end{equation}
	
	\begin{equation*}
	\text{subject to } x \in \Omega
	\end{equation*}

	\noindent where $j =\{1, 2, ... , m\}$ is the dimension of the objective space, $f_j$ is the objective function, $\omega$ is the decision (variable) space and $x$ is the variable to be optimised. The ideal point $z_j$ represents the best objective function value found so far for the $j^{th}$ objective. \\
	
	
	
	
	
	
	
	\begin{algorithm}[t]
		\caption{MOEA/D}\label{canon_alg}
		\begin{algorithmic}[1]
			\State $\Lambda = \{\lambda^{i_1}, \lambda^{i_2}, ..., \lambda^{i_N}\} \gets \text{Generate weight vectors}$
			\State $P = \{x_1, x_2, ..., x_N\} \gets \text{Initialize population}$
			\ForEach {$ \lambda^{i} \in \Lambda$} 
			\State $B(i) = \{i_1, i_2, ..., i_T\} \gets \text{Define reference table}$
			\EndFor
			\Repeat
			\ForEach {$ i \in \{1, 2, ..., N\} $} 
			\State $k, l \gets \text{Return random parent indices from B(i)}$
			\State $y \gets \text{Create child program from } x^k \text{ and } x^l$
			\State $P = Update(y, i)$
			\EndFor
			\State $EP\gets \text{Update based on P, remove non-dominated solutions}$ 
			\Until{ Stopping criteria is met}
			\State \Return {The non-dominated solutions from EP}
		\end{algorithmic}
	\end{algorithm}

	The steps involved in the algorithm are given below in more detail and pseudocode demonstrating the implementation of these steps are provided in Algorithm ~\ref{canon_alg}. \\
	
	\noindent\textbf{Step 1:} Initialisation. \\
	
	\noindent Step 1.1) Initialise external population archive $EP = \emptyset$. Initialise population $P$ with individuals $x_1, x_2...x_N$  and weight vectors $\Lambda$ (Lines 1-2 in Algorithm ~\ref{canon_alg}).\\
	
	\noindent Step 1.2) Set $B(i) = \{i_1, i_2, ...  , i_T \}$ for each $i = \{1, 2, ... ,N\}$, where $B(i)$ is a reference table of indices indicating which neighbouring subproblems are closest to the current subproblem such that $\lambda^{i_1} , \lambda^{i_2} , ..., \lambda^{i_T}$ are the $T$ closest weight vectors to $i$, determined by their Euclidean distance (Lines 3-5 in Algorithm ~\ref{canon_alg}).\\
	
	\noindent Step 1.3) Evaluate the initial fitness values $F(x_i)$.\\
	
	\noindent\textbf{Step 2:} Performing genetic operations and searching for solutions.\\
	
	\noindent Step 2.1) Select two indices $k$ and $l$ at random from the neighbourhood reference table $B(i)$. New offspring $y$ are created from parents $x_k$ and $x_l$ by performing genetic
	operations (Lines 8 - 9 in Algorithm ~\ref{canon_alg}).\\
	
	\noindent Step 2.2) Update $z$ such that for each $j = \{1, 2, ... ,m\}$ if $z_j < f_j(y)$, then set $z_j = f_j(y)$. The inequality is reversed when $F(x)$ is minimised. \\
	
	
	\noindent Step 2.3) Update the neighbouring solutions for the $j^{th}$ case such that $j \in B(i) $, if $g(y | \lambda^j, z) \geq g(x^j | \lambda^j, z)$, then let $x^j = y$ and calculate new fitness $F(y)$.\\
	
	\noindent\textbf{Step 3:} Filling the external population archive and algorithm termination.\\
	
	\noindent Step 3.1) The external population archive EP is updated with non-dominated solutions from the current generation, removing dominated solutions from the archive after updating. After the stopping criteria is satisfied output the non-dominated solutions of EP; otherwise, Step 2 is repeated (Lines 12 - 14 in Algorithm ~\ref{canon_alg}). MOEA/D is also one of the most used EMO approaches, and continues to be employed in various areas~\cite{DBLP:conf/cec/StapletonG21}.\\

	
	\section{Proposed Approach}
	\label{sec:proposed}
	\subsection{Trajectory Prediction and Objective Optimisation}

	In order to effectively carry out the task of trajectory prediction, the input data needs to be considered first. In order to generate the data the GridSim simulator is used, which emulates an ego vehicle travelling along a highway scenario and captures sequenced occupancy grids~\cite{gridsimCode,trasnea2019gridsim}. An occupancy grid is an image of a birds eye view of the ego vehicle and its surrounding environment, where the image is composed of three distinct colour coded regions: i) The area in which the vehicle is free to travel, ii) the boundary of objects and edges of the roadside and iii) the unknown regions that the ego-vehicle sensors are unable to pick up. From this information, a sequence is defined as consisting of $\tau$ occupancy grid images as $X^{<t-\tau>}$ at time $t$. Using $X^{<t-\tau>}$, the aim is to predict future trajectory positions,

	\begin{equation}
	\label{eqn:position}
	Y^{<t+\tau>} = \{x_0, y_0,\  x_1, y_1, \  x_2, y_2,\  ... , \ x_\tau , y_\tau\}
	\end{equation}
	
	\noindent where $x$ and $y$ represent the position of the ego vehicle. Recent works, including~\cite{buhet2020plop,mersch2021maneuver,chandra2020forecasting,Mo2020InteractionAwareTP},  indicated that trajectory predictions should be ranging from one to five seconds. When capturing data, the sample time for each image was at a rate of every 0.2 to 0.3 seconds, as such to have worthwhile predictions a $\tau$ value of 8 is selected, falling into approximately the two second range for future predictions. 
	
	
	A table demonstrating the prediction time window for future trajectories and performance is given in Table \ref{tab:comparison}. It is important to note that a true comparison cannot be made due to the different datasets, as well as different criteria for each work, however this shows that our results are within the correct range for state-of-the-art results.

	\begin{table}
		\small
		\centering
		\caption{Performance of state-of-the-art approaches using different datasets and different scenarios. Metric results from other authors represent a two second time frame for future intervals for RMSE. Results using different metric or time frames are not reported.}
		\label{tab:comparison}
		\resizebox{0.99\columnwidth}{!}{ 
			\begin{tabular}{|c|c|c|c|c|c|}
				\hline
				\textit{Author} & \textit{Year} & \textit{Method} & \textit{Dataset} & \textit{Time} & \textit{Performance}\\
				\hline
				Buhet et al.~\cite{buhet2020plop} & 2019 & PLOP & nuScenes~\cite{caesar2020nuscenes} & 4s & - \\
				Mo et al.~\cite{Mo2020InteractionAwareTP} & 2020 & CNN-LSTM & NGSIM~\cite{colyar2007us}   & 1s - 5s & 0.96 \\
				Mersch et al.~\cite{mersch2021maneuver} & 2021 & Spatio-temporal CNN  & NGSIM~\cite{colyar2007us}  & 5s & 1.17 \\
				Mersch et al.~\cite{mersch2021maneuver} & 2021 & Spatio-temporal CNN & highD~\cite{krajewski2018highd} & 5s & 0.21 \\
				Chandra et al.~\cite{chandra2020forecasting} & 2020 & S1+S2 & Apolloscape~\cite{huang2018apolloscape} & 3s & - \\
				Chandra et al.~\cite{chandra2020forecasting} & 2020 & S1+S2 & Argoverse~\cite{chang2019argoverse} & 5s & - \\
				Chandra et al.~\cite{chandra2020forecasting} & 2020 & S1+S2 & Lyft~\cite{kesten2019lyft} & 5s & - \\
				Chandra et al.~\cite{chandra2020forecasting} & 2020 & S1+S2 & NGSIM~\cite{colyar2007us}  & 5s & - \\
				Grigorescu~\cite{grigorescu2019ieee} & 2019 & NeuroTrajectory & GridSim~\cite{trasnea2019gridsim} & 1-2s & 0.90\\
				Galv\'{a}n and Stapleton & 2023 & Proposed method & GridSim~\cite{trasnea2019gridsim} & 2s & 0.99 \\
				\hline
			\end{tabular}
		}
	\end{table}
	
	A summary of the objectives investigated is given below. The first three objectives are related to well-defined objectives for trajectory prediction~\cite{grigorescu2019ieee}. These three objectives, denoted as l$_1$, l$_2$ and l$_3$, aid in either passenger comfort or driving experience:
	
	\begin{itemize}
		\item l$_1$ is the distance feedback which is to be minimised. By minimising the distance feedback the aim is to reduce the local travel path of the ego vehicle.
		\item l$_2$ is the lateral velocity which is also to be minimised. Here, the aim is to reduce sudden or rapid movements, which could be dangerous or cause significant passenger discomfort.
		\item l$_3$ is the longitudinal velocity which is to be maximised, helping to shorten the overall travel time of the ego vehicle.
	\end{itemize}
	
	\noindent The fourth and fifth objectives are related to performance of the ego vehicle and can be considered loss objectives:
	
	\begin{itemize}
		\item RMSE is the root mean square error and is to be minimised.
		\item SignLoss is a custom loss metric designed around vehicle behaviour.
	\end{itemize}
	
	Next, the five objectives will be discussed more explicitly. The distance feedback l$_1^{<t+\tau>}$ measures the distance between the current position of the ego vehicle $P_{ego}$ and last position in the sequence $P_{dest}$ and is expressed in Equation~\ref{eqn:l1}

	\begin{equation}
	\label{eqn:l1}
	l_1^{<t+\tau>} = \sum\limits_{i=1}^{\tau}\vert\vert P_{ego}^{<t+i>} - P_{dest}^{<t+\tau>}\vert\vert_{2}^{2}
	\end{equation}

	The lateral velocity l$_2$ is calculated from the angular velocity of the ego vehicle $v_{\delta}$, as expressed in Equations~\ref{eqn:l2} and~\ref{eqn:angvel}. Some well-known works such as~\cite{grigorescu2019ieee}, computes the lateral velocity as a function of the steering velocity, but since the GridSim simulator does not natively export this, it has instead been calculated it using the angular velocity $v_{\delta}$.
	
	\begin{equation}
	\label{eqn:l2}
	l_2^{<t+\tau>} = \sum\limits_{i=1}^{\tau} v_{\delta}^{<t+i>} 
	\end{equation}  
	
	\begin{equation}
	\label{eqn:angvel}
	\small{v_{\delta}=\frac{\atantwo(x_{i+1}-x_i,y_{i+1}-y_i)-\atantwo(x_i-x_{i-1},y_i-y_{i-1})}{t_i-t_{i-1}}}
	\end{equation}
	
	The longitudinal velocity l$_3^{<t+\tau>}$ is calculated as the component of the velocity in the $y$-direction, as expressed in Equation~\ref{eqn:l3}. Lower and upper bounds have been set on the velocity of $v_{min}$ and $v_{max}$. The GridSim simulator was updated to incorporate these constraints, where the ego vehicle travelled at velocities within the range of 80 km to 130 km at all times. The optimiser requires the objectives be minimised, as such $v_{max}$ velocity is taken away from l$_3$ during the optimisation process.
	
	
	\begin{equation}
	\label{eqn:l3}
	l_3^{<t+\tau>} = \sum\limits_{i=1}^{\tau} v_{f}^{<t+i>} \in [v_{min}, v_{max} ]
	\end{equation}
	
	The root mean squared error l$_4$ (RMSE) is calculated as the Euclidean distance between the predicted position $\hat{P}_{ego}$ of the ego vehicle and the actual position of the ego vehicle $P_{ego}$, for a trajectory at a given time-step as expressed in Equation~\ref{eqn:rmse},
	
	
	\begin{equation}
	\label{eqn:rmse}
	RMSE = \sum_{i=1}^n \frac{\sqrt{(\hat{P}_{ego} - P_{ego})^2}}{n}
	\end{equation}
	
	In addition to standard loss objectives, such as RMSE, a loss function that penalises asymmetric trajectories is also tested. An asymmetric trajectory can be categorised as a trajectory that consistently veers in a particular lateral direction, i.e., constantly veering left or right. When the trajectory data is unscaled, it is possible to incorporate the sign as an indication of the direction (negative $x$ for left and positive $x$ for right). Furthermore, trajectories are penalised that go straight-ahead by taking the absolute value of the predicted away from the real trajectory. These two criteria are combined in Equation~\ref{eqn:sign_loss} (l$_5$),
	
	\begin{equation}
	\label{eqn:sign_loss}
	Sign Loss = \frac{\sum_{i=1}^n \frac{\sqrt{(\|\hat{X}_{ego}\| - \|X_{ego}\|)^2}}{n} }{I(sign(\hat{X}) == sign(X))}
	\end{equation}
	
	A notable aspect of Equation~\ref{eqn:sign_loss} is that it is non-differentiable, as the sign function itself is non-differentiable at 0. In other words, it could not be used to train a network based on traditional gradient descent. It may be of interest, depending on the problem domain, to further investigate the use of non-differentiable functions in evolutionary algorithms, as they do not require functions to be differentiable.

	\subsection{DNNs Architecture: Convolutional Neural Network and Long-short Term Memory Network}

\begin{figure*}[t!]
		\centering
		
		\includegraphics[width=\linewidth]{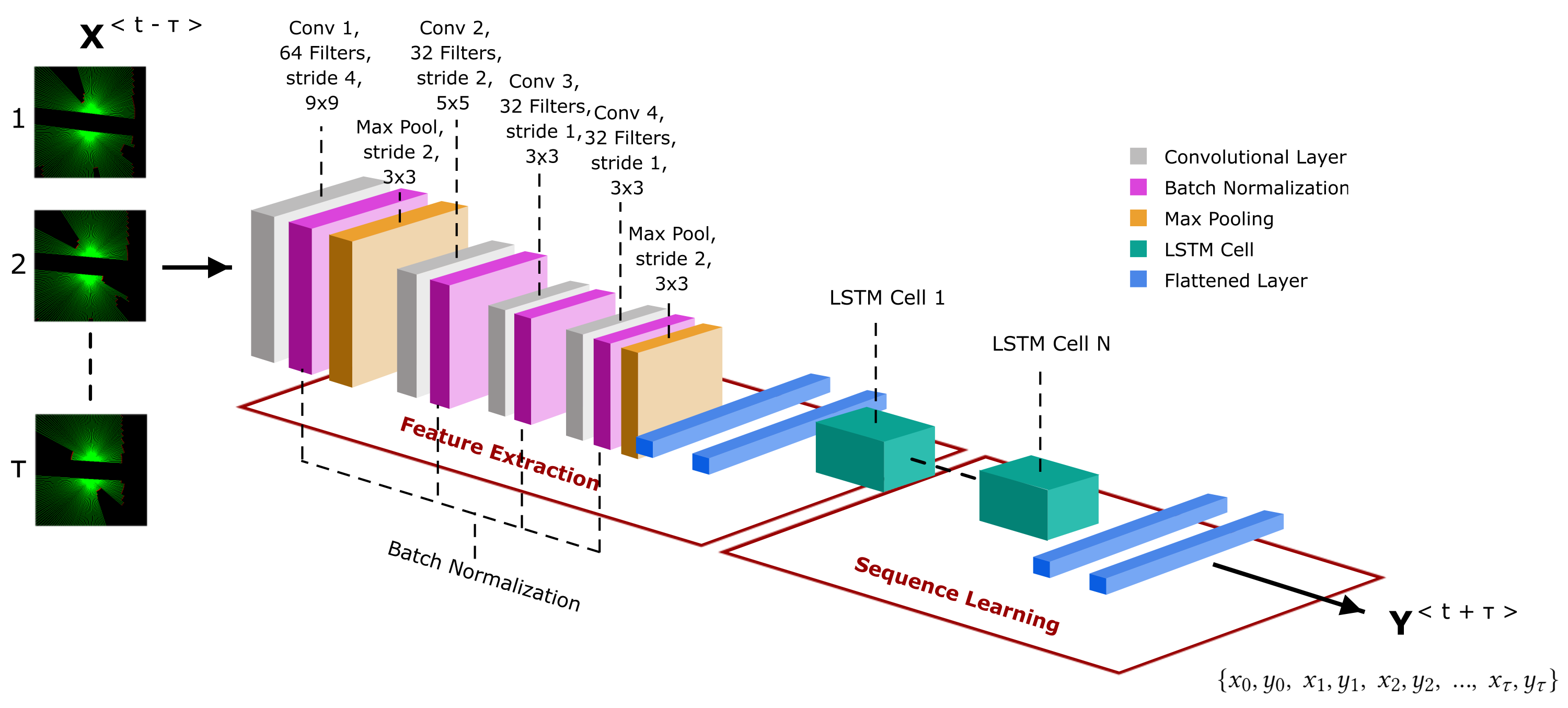}
		\caption{Diagram of network topology demonstrating two main sections of the network. i) the CNN, responsible for extraction feature information from the sequence of images, and ii) LSTM, responsible for predicting future trajectories.} 
		\label{fig:network}
	\end{figure*}

	Figure~\ref{fig:network} shows the network architecture proposed in this work. It describes both the input and output data as well as various aspects of the network design. The network is composed of two parts: i) the Convolutional Neural Network (CNN), which takes the sequence of $\tau$ images as input. The CNN is responsible for extracting important feature information from these images and ii) the Long-Short Term Memory Network (LSTM), which predicts the trajectories based on the output of the CNN. The LSTM is responsible for sequence learning. The latter network can retain or memorise long term dependencies for the sequence data, thus making it suitable for trajectory prediction. On the left of Figure~\ref{fig:network}, sequences of images of size $\tau$ are fed as input into the CNN. On the bottom right is the outputted predicted trajectory.
	
	A number of fixed hyperparameters are selected, primarily for the CNN section of the network. These fixed parameters are derived from well-known approaches including the work of Grigorescu at al. ~\cite{neurotrajCode}. These hyperparameters include kernel sizes for the CNN layers, number of filters and the type of activation function to name a few. For the non-fixed or evolvable hyperparameters, the ranges were largely changed relative to well-known methods~\cite{neurotrajCode}. In particular the range of possible values as well as additional hyperparameters. The motivation for automatically changing these values are:
	\begin{itemize}
		
		\item The LSTM scales poorly when increasing $\tau$ values with respect to the range of values for the hyperparameters chosen. In particular, the batch size and number of hidden unit hyperparameters were particularly sensitive. To counteract this issue, different batch sizes and various numbers of hidden units were made available when evolving.
		\item To help generalise for varying $\tau$ values, momentum was added to the batch normalisation. As such, this was also allowed to evolve.
		\item All of the flattened layers are available to evolve, along with an additional dropout parameter.
		
	\end{itemize}

	Table~\ref{tab:hyperparameters} lists the evolvable hyperparameters for each network. From left to right, the first column denotes the locus point, the second column specifies the gene at that locus and the third column denotes the respective list of possible alleles for that gene.
	The allele values are represented by nominal variables (loss function and optimiser) and numeric variables (all other genes). Barring LSTM Flattened Layers 1 and 2, the allele values for the numeric values have even intervals, for example, the Batch Size increments by 25 for each value. 
	
	Next, the genes are summarised. Batch Size determines the sample size for training each network. Epochs represent the number of forward and backward passes through the network for a training batch. Momentum is a parameter relating to the batch normalisation and helps alleviate noise in gradient updates.
	Two regression-based loss functions were selected which compare the true and predicted trajectories of the training data. These are the mean squared error (MSE) and log cosh (LC). The latter is similar to MSE loss, but is less sensitive to infrequent outliers~\cite{Moshagen2021FindingHD}.
	
	There are seven possible optimisers to select from: Root Mean Squared Propagation (RMSprop)~\cite{rmsprop}, Adaptive moment estimation (Adam) and its variant AdaMax~\cite{adam}, Nesterov accelerated adaptive moment estimation (NAdam)~\cite{nadam}, Stochastic Gradient Descent (SGD)~\cite{lecun2012efficient}, Adaptive Gradient Descent (AdaGrad) ~\cite{adagrad} and an approach derived from this using an adaptive learning rate Adadelta~\cite{adadelta}. A general overview of each of these optimisers and discussion of their properties is summarised in a survey by Ruder~\cite{Ruder2016AnOO}. 
	
	LSTM Cells represent the number of LSTM cells to be stacked, with LSTM dropout referring to a dropout layer occurring after each LSTM cell and Hidden Units referring to the dimensionality of the hidden state for each LSTM cell. CNNs Flattened 1 and 2 represent the size (number of nodes) of fully connected layers occurring after the CNN model. Similarly, LSTMs Flattened 1 and 2 represent the same occurring after the LSTM model. Finally, the Flattened Dropout value is a dropout value shared across all four flattened layers in the network.

	\begin{table}
		\small
		\centering
		\caption{Evolvable hyperparameters for the deep neural network used in this work.}
		\label{tab:hyperparameters}
		\begin{tabular}{|c|c|c|}
			\hline
			\textit{Locus} & \textit{Gene} & \textit{Set of possible alleles}\\
			\hline
			1 & Batch Size & \{\ 50, 75, 100, 125\ \} \\
			2 & Epochs & \{\ 10, 20, 30, 40, 50\ \} \\
			3 & Momentum & \{\ 0.8, 0.85, 0.9, 0.95\ \} \\
			4 & Loss Function & \{\ MSE, Log Cosh\ \} \\
			5 & Optimiser & \{\ RMSprop, NAdam, SGD,  \\ &&  AdaGrad, Adadelta, Adam, AdaMax\ \} \\
			6 & LSTM Cells & \{\ 1, 2, 3, 4\ \} \\
			7 & LSTM Dropout & \{\ 0.2, 0.25, 0.3, 0.35, 0.4, 0.5\ \} \\
			8 & Hidden Units & \{\ 100, 125, 150, 175, 200, 225, 250\ \} \\
			9 & CNN Flattened 1 & \{\ 256, 512, 768, 1024\ \} \\
			10 & CNN Flattened 2 & \{\ 256, 512, 768, 1024\ \} \\
			11 & LSTM Flattened 1 & \{\ 64, 128, 256, 512\ \} \\																			
			12 & LSTM  Flattened 2 & \{\ 64, 128, 256, 512\ \} \\
			13 & Flattened Dropout & \{\ 0.05, 0.1, 0.15, 0.2, 0.25\ \} \\
			\hline
		\end{tabular}
	\end{table}
	
	\newpage
	
	\section{Experimental Setup}
	\label{sec:experimental}
	
	\subsection{Training, Validation and Test Data Sets}
	\label{sec:datasets}
	
\begin{figure}
		\centering
		\includegraphics[width=0.7\linewidth]{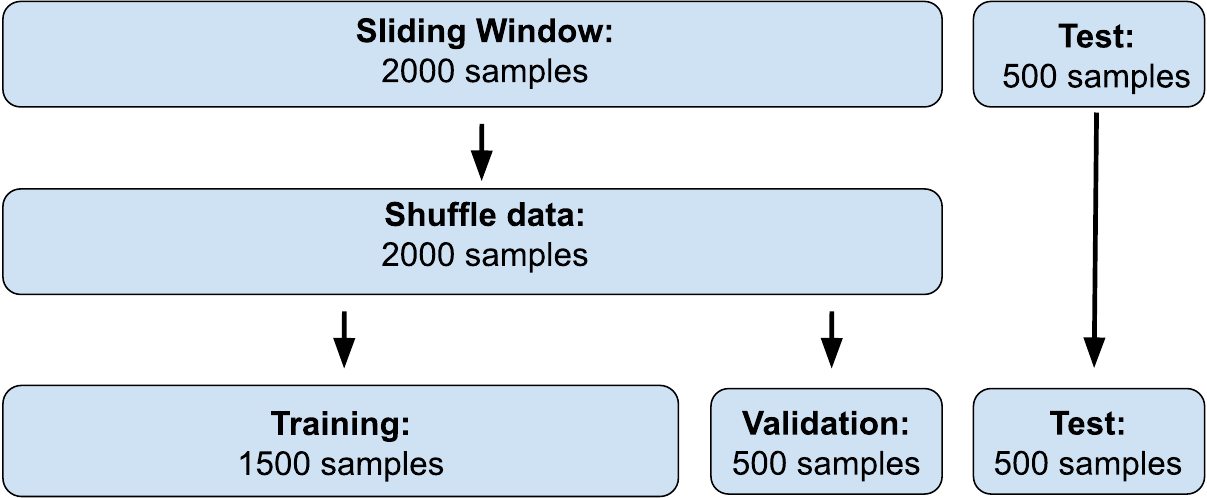}
		\caption{Diagram of data split. Sliding window and shuffle used on the training and validation sets. The test set is withheld.}
		\label{fig:data_split}
	\end{figure}

	Each instance of the data consists of a sequence of images of $\tau$ = 8 images in length, 128 x 128 pixels, with three channels of the RGB colour of each image. For the training set, there are 1500 instances of each sequence, leading to an input array of [1500, 8, 128, 128, 3] in shape, which is fed into the network. There are 500 instances for both validation and test data sets leading to a split with 0.6 : 0.2: 0.2 ratio between each data set (validation and test both have an array shape of [500, 8, 128, 128, 3]). The training and validation sets were generated using a sliding window approach to combine sequences and subsequently were shuffled prior to being split. A diagram demonstrating this data splitting technique is depicted in Figure~\ref{fig:data_split}.
	Over 30 km of data was captured using a highway scenario in the GridSim simulator~\cite{trasnea2019gridsim}. The data was cleaned to remove any near miss or dangerous manoeuvres with several seconds before and after each event being also removed to reduce any potential bias. During the experimentation, all of the objectives were calculated using the validation test. The test set is used for analysis as shown in Section~\ref{sec:results}.

	\subsection{Intensive Independent Runs and Parameters for NSGA-II and MOEA/D}
	\label{sec:runs}
	The experiments were conducted using a generational approach.
	The mutation operation works by selecting another allele value from the possible set of alleles for a given gene (see right-hand side of Table~\ref{tab:hyperparameters}). Exploitation is achieved through the use of single point crossover. This type of crossover allows one to keep the order of functions, instructions, etc., in the chromosome necessary to evolve the network for the task at hand (see left-hand side of Table~\ref{tab:hyperparameters}).
	
	A single run to report results is the norm in the deep learning community~\cite{galvan2020neuroevolution}. In this work, a leap forward is made by performing 156 independent runs (12 runs x 13 experiments), allowing for robust analysis of the results. This was done over 75 GPU days. This is the result of considering four key elements that shed light on EMO in \NEOT: i) degree of conflict among the objectives, ii) the use of highly and non-highly conflicting objectives, iii) the use of two different ground truth metrics, and iv) the use of two well-known EMO approaches.
	
	The first of these key elements is tested \textit{a priori} and is covered in more detail in Section \ref{sec:conflict}. The aforementioned key elements, two to four are handled trough three batches of experiments, totalling 13 experiments in all and are primarily covered in Sections \ref{sec:valid} to \ref{sec:2obj}. To summarise, Batch 1 of experiments investigates the use of loss functions in the presence of conflicting and non-conflicting objectives. In Batch 1, a single framework NSGA-II is used, where experiments in this batch are run for 20 generations with a population size of 25.
	
	Following on, Batch 2 investigates the role of two separate frameworks, NSGA-II and MOEA/D, selected for their distinct properties in terms of evolutionary search, namely, that one approach uses a weight mechanism, while the other uses the dominance relationships of solutions to search for Pareto optimality. Again, the experiments are testing these frameworks in the presence of conflicting and non-conflicting objectives. Experiments in this batch are run for 15 generations and with a population size of 45. In Batch 3, the analysis of the effects of conflicting and non-conflicting objectives is strengthened by reducing the problem to a bi-objective problem. In this work, the bi-objective experiments are only designed to confirm our observations from the three objective case, in practise it may be of interest to have three or more objectives.
	
	The computational complexity for the non-dominated sorting algorithm used within NSGA-II is $O(MN^2)$ where $N$ is the population size and $M$ is the number of objective functions~\cite{996017}. For MOEA/D the computational complexity is $O(MNT)$ where $T$ is the neighbourhood size, as such the computational complexity of MOEA/D is less than NSGA-II~\cite{4358754}. 
 
Furthermore, the total number of parameters for the most computational expensive CNN and LSTM model is 4,345,726 parameters. The number of evaluations required for each EMO framework, the expense of running each network and the number experiments all contribute to the overall cost, resulting 75 GPU days to complete, using Nvidia Tesla V100 16GB GPUs.
	
	
	
	Permutation tests are used for non-parametric comparison of the various models, the justification of which is i) no assumptions are made about the underlying distribution, and ii) permutation tests are appropriate tests to use when the sample size is small (12 runs per experiment are used in these experiments)~\cite{higgins2004introduction}. In addition to permutation tests we also performed Wilcoxon tests. In the case of the permutation test the null hypothesis is that the average values for each run, between two sets of experiments, belong to the same distribution, similarly the Wilcoxon tests whether there is no statistical difference between the two sets of experiments. Each test, while having different P-values, drew the same conclusions, as such when discussing whether a result is statistically different or not, it is in relation to both tests. The null hypothesis is rejected at $\alpha$ = 0.05, however, multiple tests are done of each group (each group is compared twice), so accounting for the Bonferroni correction this value is lowered to 0.025.

	
	

	\begin{table}
		\centering
		\caption{Summary of parameters used by NSGA-II and MOEA/D.}
		\resizebox{0.7\columnwidth}{!}{ 
			\small\begin{tabular}{|l|r|} \hline 
				\emph{Parameter} &
				\emph{Value} \\ \hline \hline
				
				Type of Crossover & Single point  \\ \hline
				Type of Mutation & Single point  \\ \hline
				Crossover Rate  & 1.0  \\ \hline
				Mutation Rate & 0.5 \\ \hline
				NSGA-II Selection & Tournament (size = 3)\\ \hline
				MOEA/D Neighborhood size & 7\\ \hline
			\end{tabular}
		}
		\label{tab:parameters}
	\end{table}

	\section{Discussion of Results}
	\label{sec:results}

	\subsection{Degree of Conflict Among Objectives}
	\label{sec:conflict}
	
	As discussed in Section~\ref{sec:proposed}, three domain-specific objectives are defined for the \NEOT\ problem: l$_1$ distance feedback, l$_2$ lateral velocity and l$_3$ longitudinal velocity, formally defined in Equations~\ref{eqn:l1} -~\ref{eqn:l3}. EMO approaches are normally used for conflicting objectives. Thus, it is necessary to test the conflicting nature of these objectives. It is well-known that positively correlated objectives represent objectives that are not in conflict, while negatively correlated objectives represent objectives that are indeed in conflict~\cite{deb2006searching}. It is important to note while it is still possible to optimise with two or more highly positive correlated objectives using EMO, the inclusion of these objectives offers no added benefit to the EMO process, as is the case when using NSGA-II and/or MOEA/D (see Section~\ref{sec:background}) and for certain EMO frameworks will even lead to a decrease in performance~\cite{10.1007/978-3-642-19893-9_12}.
	
	The Spearman rank-order coefficient is used to measure the rank correlation between two variables, i.e., demonstrating whether two variable can be represented using a monotonic function~\cite{zheng2019towards}. The motivation for choosing Spearman's rank-order correlation is that one cannot assume normality of blue objectives and this assumption is not a requirement of the test. The Spearman rank-order coefficient $r_s$ is calculated in Equation \ref{eqn:spearmanr} as follows:
	
	\begin{equation}
	\label{eqn:spearmanr}
	r_s = \frac{cov(R(X_1), R(X_2))}{\sigma_{R(x_1)}\sigma_{R(x_2)}}
	\end{equation}
	
	\noindent where $X_1$ and $X_2$ are the raw scores taken from two objectives. These scores are converted to ranks (i.e., $R(x_1)$, $R(x_2)$), such that the covariance $cov(R(X_1), R(X_2))$ can be calculated. $\sigma_{R(x_1)}$ and $\sigma_{R(x_2)}$ are the standard deviations of the rank variables. The values yield by this coefficient are given between -1 and +1, where a value close to -1 would indicate a highly negative correlation (objectives in conflict). Similarly, a value close to +1 would indicate a highly positive correlation (objectives not in conflict). A value of 0 would indicate no correlation.

	From Table~\ref{tab:spearmanr}, one can observe that the following pairs of objectives have a negative correlation: (l$_1$, l$_3$) (distance feedback, longitudinal velocity, respectively), (l$_2$, l$_3$) (lateral velocity for l$_2$). This negative correlation indicates that these two pairs of objectives are in conflict, although the latter pair of objectives (l$_2$, l$_3$) has a lower degree of conflict compared to the former pair of objectives (l$_1$, l$_3$). Finally, (l$_1$, l$_2$) have a positive correlation indicating that these two objectives are not in conflict. For the majority of experiments, conflicting objectives (l$_2$, l$_3$) or (l$_1$, l$_3$) are considered, along with other objectives such as RMSE or SignLoss (see Section~\ref{sec:background}). 
	Since, the RMSE and SignLoss objectives cannot be calculated \textit{a priori} we do not perform a correlation test on these objectives.
	
	\begin{table}
		\small
		\caption{Spearman rank-order correlation for distance feedback l$_1$, lateral velocity l$_2$, and longitudinal velocity l$_3$. Table inspired from~\cite{stapleton2022neuroevolutionary}.}
		\begin{tabular}{|l|c|r|r|}
			
			\hline
			
			\textit{Objectives} & \textit{Degree of Conflict}& \textit{Coefficient} & \textit{P - value} \\
			\hline

			l$_1$ \& l$_3$ & High &-0.8715 & 0.0 \\
			l$_2$ \& l$_3$ & Low &-0.1430 & 2.6771e-08 \\
			l$_1$ \& l$_2$ & Non-conflicting &0.1526 & 2.8454e-09 \\
			
			\hline 
		\end{tabular}
		\centering
		\label{tab:spearmanr}
	\end{table}
	
	\subsection{Valid Models}
	\label{sec:valid}
	Using performance metrics alone such as RMSE and SignLoss, defined in Equations~\ref{eqn:rmse} and~\ref{eqn:sign_loss}, are not enough to accurately analyse the ego-vehicle behaviour within the given driving environment. For example, they cannot rule out dangerous driving behaviours. Thus, the quality of predicted models are defined in terms of spread, symmetry (or asymmetry) and final position of the ego-vehicle. This quality is then used to quantify the number of valid models that are produced by an EMO approach at the end of each independent run.
	
\begin{figure}
		\small
		\centering
		\textbf{Real and predicted positional coordinates for Ego Vehicle}\par\medskip
		\includegraphics[width=0.99\linewidth]{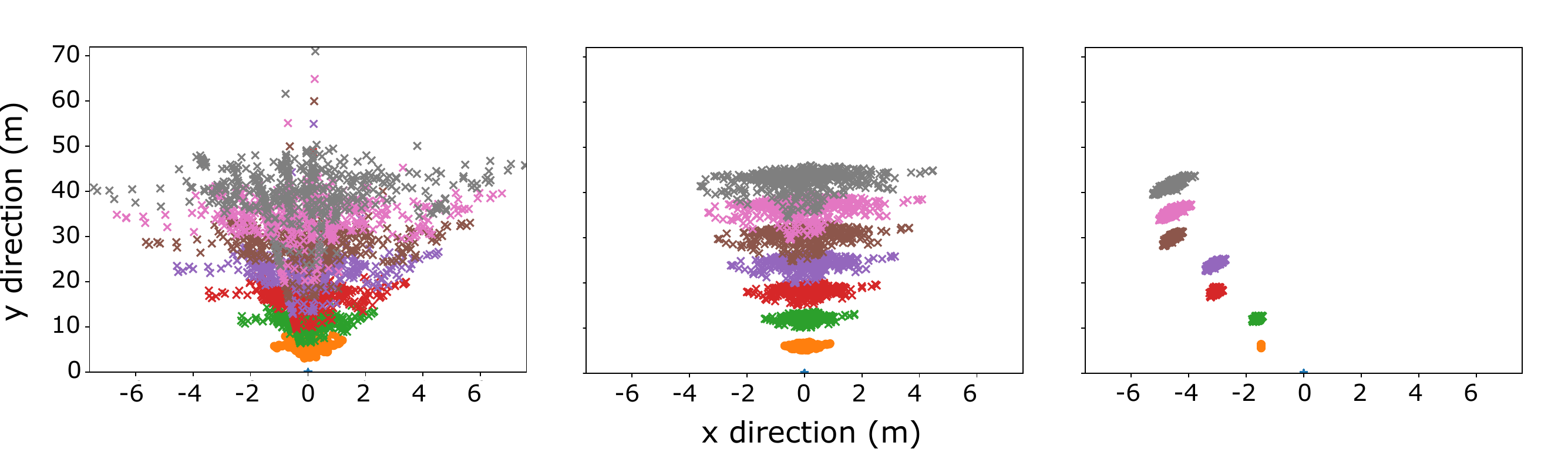}
		\caption{Spread of $x$ and $y$ positional values for real and predicted test data given in meters. Positional coordinates have been colour coded with a unique colour at each time step as a visual aid. Left figure shows the real position coordinates, centre figure shows correctly predicted positional coordinates (RMSE, l$_2$ and l$_3$) and right figure shows incorrectly predicted positional coordinates (l$_1$, l$_2$ and l$_3$).}.
		\label{fig:combined3}	
	\end{figure}

	A valid model is one that has a spread of more than $\pm$ 2m, that does not have a notable asymmetry in the predicted trajectory and which has a final position of more than 40m in the $y$-direction. The justification for these values is i) typically, a lane change would correspond to $\pm$ 2m spread, if all values fall within this range, this would indicate the ego vehicle does not readily change lanes (or only changes lanes at a slow rate), ii) asymmetry in the predicted trajectory indicates that the vehicle will consistently veer in one direction over time, and iii) that given the approximate sample rate of the images (8 images of positions every 2s), 20m/s corresponds roughly to the lower bound of the longitudinal velocity at 80km/hr (72km/hr to be precise), therefore models going above this velocity on average should be preferenced. Figure~\ref{fig:combined3} shows real positional coordinates (left), correct predicted positional coordinates using NSGA-II and RMSE, l$_2$ and l$_3$ as objectives (centre) and incorrect predicted positional coordinates (right) using NSGA-II and objectives l$_1$, l$_2$ and l$_3$.
	
	
	\subsection{Understanding the Impact of RMSE and SignLoss Measures in \NEOT\ using EMO}
	\label{sec:results3}
	
	A systematic approach is adopted, where the focus is on key elements that highlight the impact that certain objectives have on \NEOT. To this end, focus is given to using either RMSE or SignLoss, defined in Section~\ref{sec:background}, along with conflicting objectives (l$_2$, l$_3$) or (l$_1$, l$_3$). See Table~\ref{tab:spearmanr} to see the degree of conflict among these pair of objectives and their descriptions. 
	
	Of interest is the number of valid models in terms of spread, symmetry and final position of the ego-vehicle, as defined in the previous section, produced when using these objectives and NSGA-II. Table~\ref{tab:spread} shows the number of valid models expressed as fraction and percentage when using four different sets of objectives. Experiments 1 and 2, use RMSE and SignLoss, respectively, along with low conflicting objectives l$_2$ and l$_3$. From these two sets, it can be observed from Table~\ref{tab:spread} that when using RMSE, the percentage of valid models is higher compared to when using SignLoss (56\%, 44\%, respectively). Experiments 3 and 4, use RMSE and SignLoss, respectively, using highly conflicting objectives l$_1$ and l$_3$. This results in getting a significantly lower percentage of valid models with 9\%, regardless of using RMSE or SignLoss. From these results, it is learnt that either RMSE or SignLoss, along with conflicting objectives, can yield valid models for the \NEOT\ task. However, RMSE tends to yield more valid models when using objectives that are relatively in conflict, such as using l$_2$ and l$_3$ (see Table~\ref{tab:spearmanr}). 
	
	An ablation study is performed using only three objectives (l$_1$, l$_2$, l$_3$) and systematically eliminate the use of any ground truth measure, namely RMSE or SignLoss, as used in this work. Experiment 5, in Table~\ref{tab:spread}, shows a small fraction of valid models obtained by NSGA-II when trying to optimise these objectives. This gives a clear indication that the use of a ground truth measure is required in EMO for \NEOT. This and the preceding experiments, now allow the rest of the experiments to progress, considering both EMO approaches, NSGA-II and MOEA/D.

	\begin{table}
		\caption{Number of valid models obtained by NSGA-II using either RMSE or SignLoss along with conflicting objectives (l$_2$, l$_3$) and (l$_1$, l$_3$). Table has been reproduced in part from ~\cite{stapleton2022neuroevolutionary}.}
		\centering
		\resizebox{1\textwidth}{!}{ 
			\begin{tabular}{l|rr|rr|r|}
				
				&
				\textit{Experiment 1} & 
				\textit{Experiment 2} &
				\textit{Experiment 3} & 
				\textit{Experiment 4} &
				\textit{Experiment 5} \\
				\hline

				
				& \textit{ ($RMSE$, $l_2$, $l_3$)}
				& \textit{ ($SignLoss$, $l_2$, $l_3$)}
				& \textit{ ($RMSE$, $l_1$, $l_3$)} 
				& \textit{ ($SignLoss$, $l_1$, $l_3$)}
				& \textit{ ($l_1$, $l_2$, $l_3$)} \\
				
				\hline
				\hline
				
				Fraction & 168/300 & 132/300 & 26/300 & 28/300 & 5/300 \\
				Percentage & 56\% & 44\% & 9\% & 9\% & 2\% \\
				\hline
				
		\end{tabular}}
		\label{tab:spread}
	\end{table}
	
	
	Table~\ref{tab:results} shows the average RMSE for the validation and test set (RMSE$_{val}$ (all), RMSE$_{test}$ (all); first and second row, respectively) across all runs, using non-dominated solutions from the final generation. The above-mentioned metrics represent all models, both valid and non-valid. These values are expected to conform with the observations from Table \ref{tab:spread} as they take into consideration all models, whether they are valid or not. In the below analysis Experiments 1 vs. 3 and Experiments 2 vs. 4 are compared, but like before this can be considered as a comparison against the low conflicting combinations (l$_2$, l$_3$) and high conflicting combinations (l$_1$, l$_3$). When comparing Batch 1 (Experiments 1 and 2) with Batch 2 (Experiments 3 and 4), it can be seen that both the RMSE$_{val}$ (all) and RMSE$_{test}$ (all) are statistically significantly lower for Batch 1 based on permutation and Wilcoxon tests (for details, see Section \ref{sec:runs}). This is notable as the lower RMSE values and hence preferable values correspond to the higher percentage of valid models found with the weakly conflicting objectives observed in Table \ref{tab:spread}. 
	
	\begin{table}[t!]
		\caption{Mean and standard deviation ($\pm$) of real trajectory vs. predicted trajectory. Metrics denoted by `all' represent both valid and non-valid models. Experiments 1 and 2 belong to Batch 1 and experiments 3 and 4 belong to batch 2. Table has been reproduced in part from ~\cite{stapleton2022neuroevolutionary}.}
		\centering
		\resizebox{1\textwidth}{!}{ 
			\begin{tabular}{l|rr|rr|rr|rr|}

				& \multicolumn{2}{c}{Experiment 1} 
				& \multicolumn{2}{c}{Experiment 2} 
				\vline
				& \multicolumn{2}{c}{Experiment 3} 
				& \multicolumn{2}{c}{Experiment 4} 
				
				\vline\\
				& \multicolumn{2}{c}{NSGA-II} 
				& \multicolumn{2}{c}{NSGA-II} 
				\vline
				& \multicolumn{2}{c}{NSGA-II} 
				& \multicolumn{2}{c}{NSGA-II} 
				
				\vline\\
				
				& \multicolumn{2}{c}{($RMSE$, $l_2$, $l_3$)} 
				& \multicolumn{2}{c}{($SignLoss$, $l_2$, $l_3$)} 
				\vline
				& \multicolumn{2}{c}{($RMSE$, $l_1$, $l_3$)} 
				& \multicolumn{2}{c}{($SignLoss$, $l_1$, $l_3$)} 
				\vline \\
				
				\hline
				\textit{metric} &     \textit{mean} &      \textit{std} &    \textit{mean} &    \textit{ std} &     \textit{mean} &     \textit{std} &     \textit{mean} &     \textit{std}  \\
				\hline
				\hline
				
				RMSE$_{val}$ (all) & 0.657 & 0.0327 & 0.728 & 0.0571 &  2.078 & 0.3517 &  2.308 & 0.2350 \\
				

				RMSE$_{test}$ (all) & 1.063 & 0.0080 & 1.049 & 0.0101 &  1.606 & 0.2047 &  1.734 & 0.1528   \\
				\hline
				RMSE$_{val}$ (valid only) & 0.538 & 0.0402 & 0.537 & 0.0552 &  0.595 & 0.2025 &  0.591 & 0.1322  \\
				RMSE$_{test}$ (valid only) & 1.070 & 0.0188 & 1.051 & 0.0238 &  1.098 & 0.0765 &  1.078 & 0.0660 \\
				\hline 
			\end{tabular}
		}
		\label{tab:results}
	\end{table}

	Next, the valid models are investigated. These are represented by RMSE$_{val}$ (valid only) and RMSE$_{test}$ (valid only) third and fourth row, respectively). These represent the average of only the non-dominated valid models, across all runs for the final generation. First, analysis RMSE$_{val}$ (valid only) values is conducted, comparing again Batch 1 and Batch 2. It is found that these values were not statistically different based on permutation and Wilcoxon tests. This can be attributed to the relatively high standard deviations of Experiments 3 and 4 ($\pm$ 0.2025, $\pm$ 0.1322, respectively). Next, the same comparison on RMSE$_{test}$ (valid only) is performed, where again it is found that the differences are not statistically significant. In summary, Batch 1 performs better than Batch 2 when both valid and non-valid models are considered, while the differences were not statistically significant for valid only models. To fully understand this behaviour row-wise comparison is subsequently done, where the effect of removing non-valid models on the performance metrics is investigated.

	

	
	Considering only Batch 1, there is an improvement between RMSE$_{val}$ (all) and RMSE$_{val}$ (valid only) (Rows 1 and 3 respectively, which represent the validation data set), however, there is no significant change between RMSE$_{test}$ (all) and RMSE$_{test}$ (valid only) (Rows 2 and 4, which represent the test data set). Now, if the same comparison for Batch 2 is done there is a substantial improvement for RMSE$_{val}$ and RMSE$_{test}$ for both the `all' and the `valid only' cases. In other words, only RMSE$_{test}$ values for Batch 1 were found to not have a statistical improvement as a result of removing valid models.
	
	This demonstrates two things. Firstly, in the case of Batch 1 (i.e., the highly conflicting combinations) when the non-valid models are removed, an improvement in performance for the validation set that does not translate to the test set is observed. This may be a result of overfitting. Another consideration is that there are more valid models in the final generation for Batch 1, and as such, the removal of non-valid models is less impactful. Secondly, there is a much larger improvement in Batch 2 when the non-valid models are removed. This demonstrates that not only do the non-valid models greatly affect performance but also that the few valid models that are retained are comparable in performance to Batch 1.

	Finally, a comparison of the SignLoss and RMSE objectives is done, where Experiments 1 vs. 2 and Experiments 3 vs. 4 are compared. When both the valid and non-valid models are considered. the RMSE objective generally outperforms the SignLoss function, barring Experiment 2 where RMSE$_{test}$ (all) was better for the SignLoss objective (1.049 $\pm$ 0.0101). For valid only models the comparison is inconclusive, in other words, the differences were not found to be statistically significant. In particular, Experiment 2 had a relatively low P-value of 0.05 for the permuation test (P-value of 0.08 for the wilcoxon test) but since adjusting for Bonferroni's correction this cannot be considered significant.
	
	

	From the above results we learn the following (bearing in mind we are considering the objectives: root mean squared error (RMSE), the sign loss (SignLoss), distance feedback (l$_1$), lateral velocity (l$_2$), and longitudinal velocity (l$_3$).):
	
	
	\begin{itemize}
		\item In particular, the inclusion of l$_1$ seems to be detrimental to overall performance in terms of valid models discovered and RMSE$_{val}$ (all) and RMSE$_{val}$ (valid only).
		\item The detrimental effects on performance of the highly conflicting objectives (l$_1$, l$_3$) are related to the heuristics ability to find valid models and not necessarily on the quality of found models. It can be seen that once non-valid models are removed, the (l$_1$, l$_3$) experiments are almost comparable to the low conflicting experiments containing (l$_2$, l$_3$).
		\item While the SignLoss offers some marginal improvement performance in terms of RMSE$_{test}$ (all) with mixed results for other metrics, it was decided to retain RMSE for future batches of experiments as it produced more valid models, at least in the case of Batch 1.
		
	\end{itemize}


	

	
	\subsection{Analysing Conflicting Objectives in \NEOT\ using NSGA-II and MOEA/D}
	\label{sec:analyse_conflict}
	
	From the previous section, it is understood that RMSE produces a higher percentage of valid models for the \NEOT\ task. It was also seen how NSGA-II was able to use RMSE along with conflicting objectives yielding good results in terms of correct trajectory predictions. Following on from this, the analysis is continued, but now using both EMO approaches: NSGA-II and MOEA/D, which have completely different underlying mechanisms. One using a Pareto dominance-based approach (NSGA-II) and the other one using a decomposition mechanism (MOEA/D). Table~\ref{tab:parameters} shows the parameters’ values for each of these two algorithms.
	
	Focus is now drawn to the number of valid models produced by either NSGA-II or MOEA/D. Table~\ref{tab:spread2} shows the fraction and percentage of valid models obtained by either NSGA-II or MOEA/D and conflicting objectives. It is interesting to see that the percentage of valid models is higher when using MOEA/D instead of NSGA-II, regardless of the pairs of conflicting objectives used. For example, when using highly conflicting objectives (l$_1$, l$_3$), MOEA/D produces 6\% more valid models compared to NSGA-II (see right-hand side of Table~\ref{tab:spread2}). The same occurs when using less conflicting objectives (l$_2$, l$_3$). In this case, it is seen that again MOEA/D produces 4\% more valid models compared to NSGA-II. Moreover, it is worth noting that the decomposition EMO approach produces many more models compared to NSGA-II. This is the result of the mechanism employed by MOEA/D, which is based on employing an archive, as discussed in Section~\ref{sec:background}.

	\begin{table}
		\small
		\caption{Analysis of trajectory spread denoting valid models in the final generation of each run.}
		\centering
		\resizebox{1\textwidth}{!}{ 
			\begin{tabular}{l|rr|rr|}

				&
				\textit{Experiment 6} & 
				\textit{Experiment 7} &
				\textit{Experiment 8} & 
				\textit{Experiment 9}  \\
				\hline
				& \textit{NSGA-II} 
				& \textit{MOEA/D} 
				& \textit{NSGA-II} 
				& \textit{MOEA/D} \\
				
				& \multicolumn{2}{c}{\textit{ ($RMSE$, $l_2$, $l_3$)}}
				\vline
				& \multicolumn{2}{c}{\textit{ ($RMSE$, $l_1$, $l_3$)}}
				\vline \\
				
				\hline
				\hline

				Fraction & 206/540 & 266/638 & 44/540 & 129/908  \\
				Percentage & 38\% & 42\% & 8\% & 14\%  \\
				\hline
				
		\end{tabular}}
		\label{tab:spread2}
	\end{table}

	Focusing now on the mean and standard deviation of real trajectory vs. predicted trajectory for the four objectives analysed in this section; the root mean squared error (RMSE), distance feedback (l$_1$), lateral velocity (l$_2$), and longitudinal velocity (l$_3$), using either NSGA-II or MOEA/D. These are shown in Table~\ref{tab:results2}. When looking at the RMSE$_{val}$ (all) and RMSE$_{test}$ (all), the difference of values between NSGA-II and MOEA/D for RMSE, l$_2$ and l$_3$ are similar (Experiments 6 and 7). However, this situation changes when using objectives that are highly in conflict: RMSE, l$_1$ and l$_3$ (Experiments 8 and 9). In this instance, it can be seen that, in general, the decomposition-based method (MOEA/D) performs better than the Pareto dominance-based approach (NSGA-II). Thus, when considering all models (valid and non-valid), the inclusion of a highly conflicting objective (l$_1$) is handled better by the MOEA/D approach. Again, looking at the valid only models, the difference in values for RMSE$_{val}$ (all) and RMSE$_{test}$ (all) are statistically insignificant for the weakly conflicting RMSE, l$_2$ and l$_3$. Interestingly, for the highly conflicting objectives the RMSE$_{test}$ (valid only) models are better for NSGA-II than MOEA/D, while in the case of RMSE$_{val}$ (valid only) the results are again better for MOEA/D. The result seems to confirm that there is a degree of overfitting occurring because even though the RMSE$_{test}$ (valid only) for MOEAD is comparatively higher, this nevertheless resulted in preferable results for the validation data set. Overall, it can be concluded that in the case of the highly conflicting objectives MOEA/D has produced significantly better results, while for the weakly conflicting objectives it is not possible to conclude whether one EMO framework is better than the other.

	\begin{table*}[t!]
		\caption{Metric mean and standard deviation (std) for all runs of a given experiment. Experiments 6 and 8 represent the experiments for NSGA-II, Experiments 6 and 8 represent the experiments for MOEA/D.}
		\centering
		\resizebox{1\textwidth}{!}{ 
			\begin{tabular}{l|rr|rr|rr|rr|}

				& \multicolumn{2}{c}{Experiment 6} 
				& \multicolumn{2}{c}{Experiment 7}
				\vline
				& \multicolumn{2}{c}{Experiment 8} 
				& \multicolumn{2}{c}{Experiment 9}
				\vline\\
				
				& \multicolumn{2}{c}{NSGA-II } 
				& \multicolumn{2}{c}{MOEA/D } 
				\vline
				& \multicolumn{2}{c}{NSGA-II } 
				& \multicolumn{2}{c}{MOEA/D } \vline\\
				
				& \multicolumn{2}{c}{($RMSE$, $l_2$, $l_3$)} 
				& \multicolumn{2}{c}{($RMSE$, $l_2$, $l_3$)} 
				\vline
				& \multicolumn{2}{c}{($RMSE$, $l_1$, $l_3$)} 
				& \multicolumn{2}{c}{($RMSE$, $l_1$, $l_3$)} 
				
				\vline\\
				\hline
				
				\textit{metric} &     \textit{mean} &      \textit{std} &    \textit{mean} &    \textit{ std} &     \textit{mean} &     \textit{std} &     \textit{mean} &     \textit{std} \\
				\hline
				\hline
				
				RMSE$_{val}$ (all)  & 0.627 & 0.0297 & 0.625 & 0.0312 &  1.757 & 0.2926 &  1.411 & 0.1364 \\

				RMSE$_{test}$ (all) & 1.065 & 0.0076 & 1.066 & 0.0084 &  1.422 & 0.1659 &  1.280 & 0.0882 \\
				\hline 
				
				RMSE$_{test}$ (valid only) & 0.534 & 0.0413 & 0.544 & 0.0283 &  0.546 & 0.0771 &  0.637 & 0.0666 \\
				
				RMSE$_{val}$ (valid only) & 1.071 & 0.0208 & 1.071 & 0.0091 &  1.053 & 0.0260 &  1.037 & 0.0242 \\
				\hline 
			\end{tabular}
		}
		\label{tab:results2}
	\end{table*}
	
\begin{figure}
		\centering
		\includegraphics[width=\linewidth]{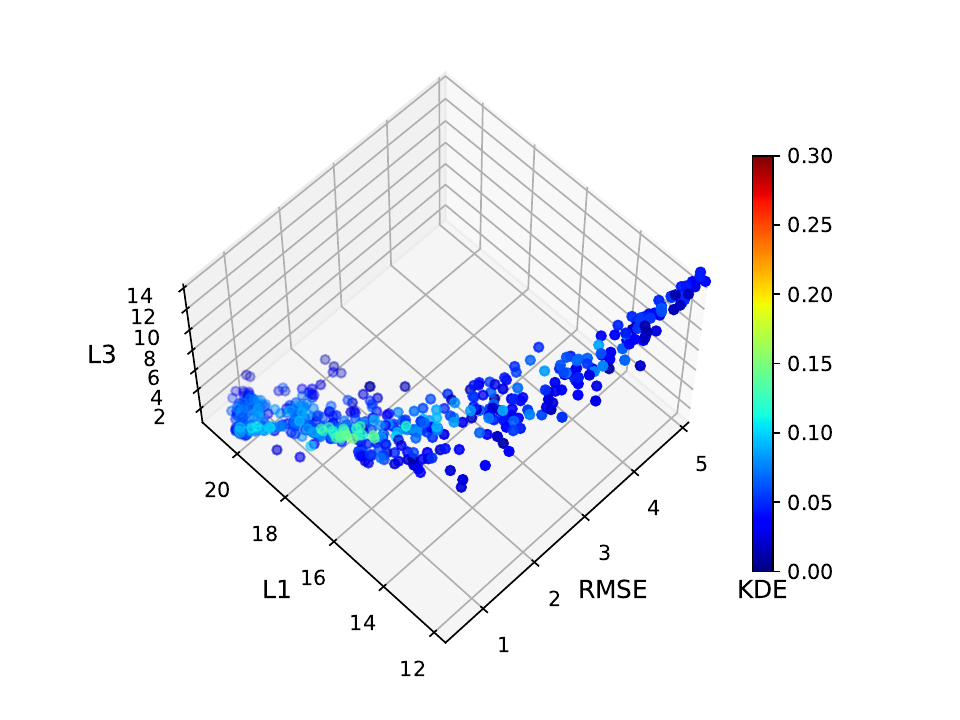}
		\caption{Plot of final generation for NSGA-II, all runs combined. Kernel density estimation (KDE) calculated independently for each run. The average mean and standard deviations for (RMSE, l$_1$, l$_3$) are (1.76 $\pm$ 0.292,\textbf{ 16.59 $\pm$ 0.460}, 6.43 $\pm$ 0.707), respectively, where the lower value compared to Figure \ref{fig:moead_front} has been highlighted in bold. Red indicates higher density of solutions, while blue indicates lower density.
		}
		\label{fig:nsgaii_front}
	\end{figure}

\begin{figure}
		\centering
		\includegraphics[width=\linewidth]{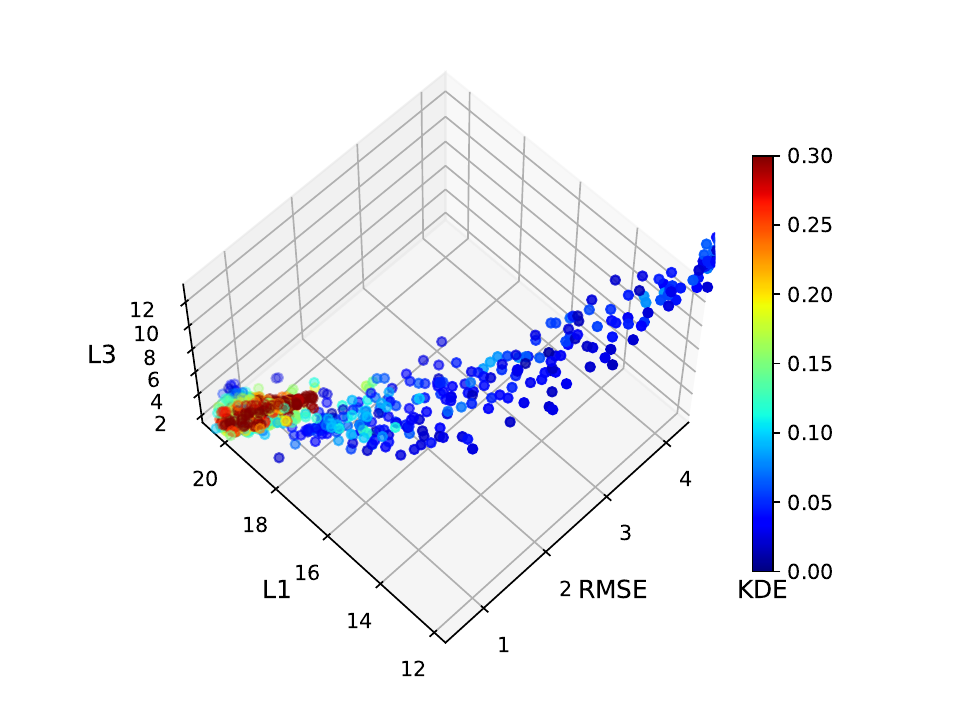}
		\caption{Plot of external population archive for MOEA/D, all runs combined. Kernel density estimation (KDE) was calculated independently for each run. The average mean and standard deviations for (RMSE, l$_1$, l$_3$) are (\textbf{1.41 $\pm$ 0.136}, 18.01 $\pm$ 0.256, \textbf{4.90 $\pm$ 0.24}), respectively, where the lower values compared to Figure \ref{fig:nsgaii_front} has been highlighted in bold. Red indicates higher density of solutions, while blue indicates lower density.}
		\label{fig:moead_front}
	\end{figure}

	Next, the Pareto front of the last generation for NSGA-II in Figure~\ref{fig:nsgaii_front} and the Pareto front of the external population archive for MOEA/D n Figure~\ref{fig:moead_front} across all runs are plotted. In Figures~\ref{fig:nsgaii_front} and ~\ref{fig:moead_front} the density of solutions is calculated using Gaussian kernel density estimation (KDE) with Scott's Rule for determining the bandwidth~\cite{scott2015multivariate}. The density is estimated independently for each run and afterwards the results of each run are combined together into each figure. The color bar to the right of each figure represents the density, where red denotes a high density and blue denotes a low density. 
	When comparing Figures~\ref{fig:nsgaii_front} and~\ref{fig:moead_front}, it can be seen that in Figure~\ref{fig:moead_front} (corresponding to MOEA/D) there is a larger density of values around the RMSE and l$_3$ axis in the bottom left of the figure, while for Figure~\ref{fig:nsgaii_front} (corresponding to NSGA-II) the density of solutions are lower and more equally spread. This analysis confirms that the MOEA/D approach is focusing more on the RMSE and l$_3$ objectives, which is in turn leading to a lower and more preferable RMSE value.

\begin{figure}
		\centering
		\includegraphics[width=0.9\linewidth]{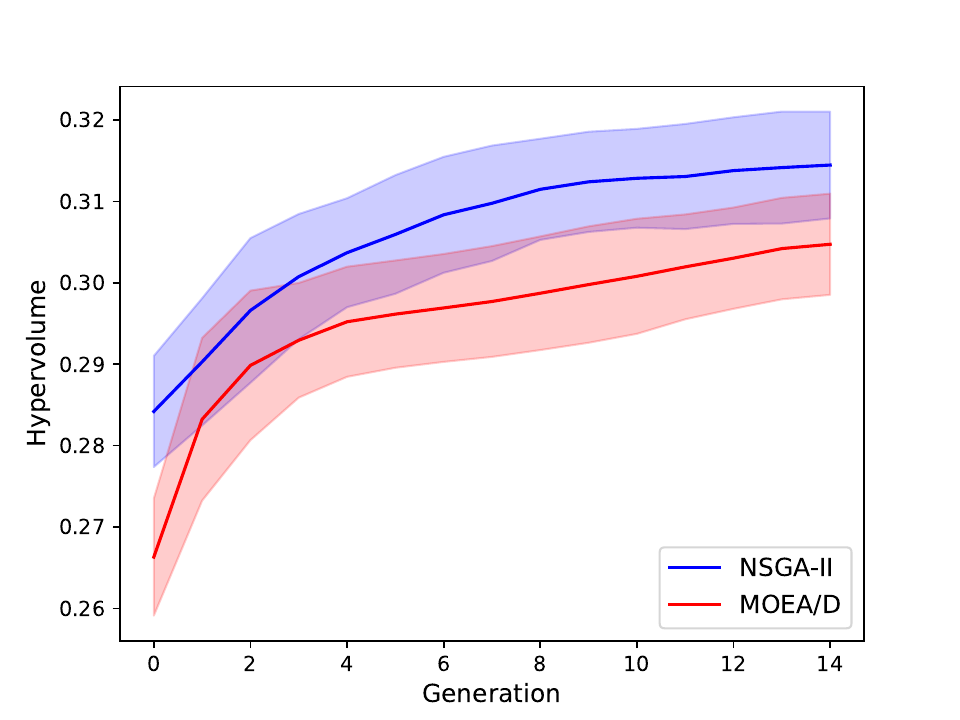}
		\caption{Plot of hypervolume vs. generation for NSGA-II and MOEA/D. In this case higher hypervolumes are preferable. }
		\label{fig:hv_comp}
	\end{figure}

	It is important to address why the MOEA/D algorithm focuses more on the RMSE and l$_3$ objectives when compared to NSGA-II. The objectives are not scaled and for the decomposition approach the weights are uniformly sampled, this in turn results in a greater density of sub-problems at the intersection of the RMSE and l$_3$ objectives. As such, for a different set of objectives and for a different problem domain therefore this improvement would not therefore be guaranteed, however, the flexibility to choose different weight initialisation schemes allows the researcher to select greater weighting towards more preferable objectives such as a loss objective like RMSE. Furthermore, in Figure~\ref{fig:hv_comp} hypervolume is plotted at each generation for both approaches for objectives (RMSE, l$_1$, l$_3$). In this case, it can be seen that at every generation NSGA-II has a higher hypervolume when compared to MOEA/D. This means that NSGA-II has better approximated the Pareto optimal front though the approach has produced less valid models. This has occurred since NSGA-II is less constrained to the scaling of objectives and hence is less sensitive to the shape of the Pareto front in objective space. As such, if one wishes to promote all objectives equally (i.e., when model performance is not tied to specific objectives) then NSGA-II is still a good approach. However, if one wishes to bias their optimisation towards one or more preferential objectives, then MOEA/D may be a better option, since they could potentially exploit the sampling of objective space by using non-uniform weighting.

	\subsection{Analysis of two-objective problem} 
	\label{sec:2obj}

	Attention is now drawn to the bi-objective experiments from Batch 3. Table \ref{tab:spread3} again shows the trajectory spread analysis, where the first row gives the valid models as a fraction the total number of models
	across all runs and the row beneath gives the percentage. When
	comparing Experiments 10 and 11 (Category 6) there is a 4\%
	increase from 38\% to 40\% when comparing NSGA-II with
	MOEA/D for (RMSE, l$_2$) objectives. For Experiments 12
	and 13 (Category 7) there was no change in percentage between
	NSGA-II with MOEA/D for (RMSE, l$_1$). Overall, in the bi-objective case, there is no notable difference in viable models between NSGA-II and MOEA/D. Looking at the objective combinations of (RMSE, l$_2$) and (RMSE, l$_1$), l$_2$ appears to be beneficial for finding viable models where ~40\% of models were useful in experiments 10 an 11. Again, the inclusion of l$_1$ was detrimental in finding valid models where ~10\% of models were useful in Experiments 12 and 13.

	In Table \ref{tab:results3} it is shown the RMSE metrics for the bi-objective cases, again comparing NSGA-II and MOEA/D. Comparing Experiments 10 and 11 where the objective combination (RMSE, l$_2$) is considered, no statistical difference was found when considering all four RMSE metrics (i.e., both the `all' and `valid only' metrics). This directly parallels the previous observation for of the weakly conflicting combinations (l$_2$, l$_3$) of the three objective case (Experiments 6 and 7), thus strengthening the analysis that the inclusion of the l$_2$ objective is beneficial.
	
	Looking at RMSE$_{val}$ (all) and RMSE$_{test}$ (all), NSGA-II would appear to have a better performance for the (RMSE, l$_1$) combination, however, these were not found to be statistically significant (P-value $>$ 0.1, in both cases for both permutation and Wilcoxon tests). This can be attributed to the relatively higher standard deviations of these results. The high level of variance can be explained by the low number of non-dominated solutions for the (RMSE, l$_1$) combination. When looking at the valid only cases, RMSE$_{val}$ (valid only) was found to be statistically lower but with a relatively large P-value ($\sim$0.02, for both permutation and Wilcoxon tests). The values for RMSE$_{test}$ (valid only) were not found to be statistically significant.
	
	
	\begin{table}
		\caption{Analysis of trajectory spread denoting valid models in the final generation of each run.}
		\centering
		\resizebox{1\textwidth}{!}{ 
			\begin{tabular}{l|rr|rr|}

				&
				\textit{Experiment 10} & 
				\textit{Experiment 11} &
				\textit{Experiment 12} & 
				\textit{Experiment 13}  \\
				\hline
				
				& \textit{NSGA-II} 
				& \textit{MOEA/D} 
				& \textit{NSGA-II} 
				& \textit{MOEA/D} \\

				& \multicolumn{2}{c}{\textit{ ($RMSE$, $l_2$)}}
				\vline
				& \multicolumn{2}{c}{\textit{ ($RMSE$, $l_1$)}}
				\vline \\
				
				\hline
				\hline
				

				Fraction & 205/540 & 135/320 & 55/540 & 20/191  \\
				Percentage & 38\% & 42\% & 10\% & 10\%  \\
				\hline
				
		\end{tabular}}
		\label{tab:spread3}
	\end{table}

	
	
	\begin{table*}[t!]
		\caption{Metric mean and standard deviation (std) for all runs of a given experiment. Experiments 10 and 12 represent the experiments for NSGA-II, Experiments 11 and 13 represent the experiments for MOEA/D.}
		\centering
		\resizebox{1\textwidth}{!}{ 
			\begin{tabular}{l|rr|rr|rr|rr|}

				& \multicolumn{2}{c}{Experiment 10} 
				& \multicolumn{2}{c}{Experiment 11}
				\vline
				& \multicolumn{2}{c}{Experiment 12} 
				& \multicolumn{2}{c}{Experiment 13}
				\vline\\
				
				& \multicolumn{2}{c}{NSGA-II } 
				& \multicolumn{2}{c}{MOEA/D } 
				\vline
				& \multicolumn{2}{c}{NSGA-II } 
				& \multicolumn{2}{c}{MOEA/D } \vline\\
				
				& \multicolumn{2}{c}{($RMSE$, $l_2$)} 
				& \multicolumn{2}{c}{($RMSE$, $l_2$)} 
				\vline
				& \multicolumn{2}{c}{($RMSE$, $l_1$)} 
				& \multicolumn{2}{c}{($RMSE$, $l_1$)} 
				\vline\\
				\hline
				
				\textit{metric} &     \textit{mean} &      \textit{std} &    \textit{mean} &    \textit{ std} &     \textit{mean} &     \textit{std} &     \textit{mean} &     \textit{std} \\
				\hline
				\hline
				
				\hline
				
				
				RMSE$_{val}$ (all) & 0.545 & 0.0317 & 0.536 & 0.0169 &  1.820 & 0.4762 &  2.124 & 0.5055 \\
				RMSE$_{test}$ (all) & 1.027 & 0.0095 & 1.025 & 0.0067 &  1.581 & 0.2839 &  1.679 & 0.3074 \\
				\hline
				
				RMSE$_{val}$ (valid only) & 0.444 & 0.0425 & 0.448 & 0.0236 &  0.449 & 0.0972 &  0.545 & 0.0751 \\
				RMSE$_{test}$ (valid only) & 1.040 & 0.0107 & 1.039 & 0.0152 &  1.007 & 0.0270 &  0.994 & 0.0365 \\
				\hline 
			\end{tabular}
		}
		\label{tab:results3}
	\end{table*}
	
	\label{sec:2obj_C}

	The analysis helps confirm the observations from Section~\ref{sec:analyse_conflict}, namely, that the lateral velocity objective l$_2$ is beneficial for finding viable models and the distance feedback l$_1$ is detrimental for finding viable models. In general, there was no major difference in performance between the NSGA-II and the MOEA/D approach in the bi-objective case. A further comparison of Tables~\ref{tab:spread2} and ~\ref{tab:spread3} reveals the bi-objective case has a much smaller number of solutions in the external archive when compared to three objective cases. For instance in the latter case there are 266/638 cases compared to the bi-objective experiment where there are 135/320 cases (considering (RMSE, l$_2$, l$_3$) and (RMSE, l$_2$)) and similarly, 129/908 cases compared to the bi-objective experiment where there are 20/191 cases (considering (RMSE, l$_1$, l$_3$) and (RMSE, l$_1$)). This means that while the exclusion of the longitudinal velocity l$_3$ does not affect the general percentages in terms of valid models, nor the general observations of the affect on performance of the other objectives (l$_1$ and l$_2$), its removal makes a fair comparison of NSGA-II vs. MOEA/D more difficult in the bi-objective case.
	
	\subsection{Our contribution to the ongoing research in EMO in Neuroevolution}
	
	Our work has made a contribution in evolutionary multi-objective optimisation in neuroevolution. Specifically, an in-depth study has been performed, using various objectives important to the area of trajectory prediction for autonomous vehicles. In this work, it has been demonstrated, what the most relevant objectives are to effectively perform this task. Specifically, distance feedback l$_1$ was found to be detrimental to evolutionary search, while longitudinal velocity l$_2$ was beneficial. This has been concluded through robust statistical analysis over a total of 156 independent runs, as opposed to a single run as normally performed by the deep learning community~\cite{galvan2020neuroevolution}. Complementing this, results have been analysed over four well-defined metrics, RMSE$_{val}$ (all), RMSE$_{test}$ (all) RMSE$_{val}$ (valid only) and RMSE$_{test}$ (valid only). Furthermore, through a series of \textit{a priori} correlation tests as well as a series of ablation studies, the number of experiments required to make conclusions were greatly reduced. It is hoped that these techniques will benefit future practitioners by reducing the computational cost of running potentially unnecessary tests.
	
	An expansive list of hyperparameters were made available to the EMO approach, as documented in Table~\ref{tab:hyperparameters}, allowing for complex and rich network designs to be tested during the neuroevolutionary search. To test and validate the approach, NSGA-II and MOEA/D are used, which are both robust and widely accepted EMO approaches. Another important contribution is to highlight the importance of objective scaling when considering which EMOs to use. In particular, the proposed recommendations are made for when it is not possible to normalise objectives, which is often the case when considering real world data. It was found that NSGA-II is less susceptible to the scaling of objectives. However, MOEA/D allows one to concentrate on specific objectives through the weight initialisation, therefore, knowing which objectives to preference beforehand can help one attain better performance.
	
	
	%
	
	%
	
	
	\section{Conclusion}
	\label{sec:conclusions}
	
	
	To date there have been few works to explore the use of Evolutionary Multi-objective optimisation to find optimal architectures in Deep Neural Networks. Using the challenging problem domain of trajectory prediction extensive analysis is undertook,  analysing the role problem-specific objectives play on the search performance. To do so,  a set of criteria are established to evaluate the quality of model architectures found in terms of the ego vehicles performance within its environment. With these quality metrics the importance of carefully selecting objectives is emphasised, since some combinations of objectives can lead to very few valid models being found. The analysis found that the inclusion of the distance feedback l$_{1}$ was considerably detrimental to finding valid models, while the lateral velocity l$_2$ was more beneficial. Furthermore, testing for correlation among the objectives a priori, along with a series of ablation studies helped greatly reduce the number of experiments that needed to be performed.
	
	A comparative study was done between two very different EMO frameworks. The analysis concluded that NSGA-II was preferential for maximising the hypervolume with non-scaled objectives. On the other hand, MOEA/D can be designed to concentrate on specific objectives, which in this case,
	led to more valid models being found during evolutionary search.
	It is hoped that drawing attention to the importance of scaling will be beneficial to other problem-domains outside of autonomous vehicle, where the normalisation of objectives is often not guaranteed. 
	
	While this work demonstrates the importance of objective scaling, one limitation is that we only compare scenarios where objective scaling cannot be guaranteed. It would be interesting to test each EMO on objectives that are scaled vs. not scaled, however to do so would likely require benchmarks outside of neurotrajectory prediction. Furthermore, we have tested our scenario in within a single context i.e., simulated highway data. Additional driving contexts as well as testing on real data would be of interest.
	
	In the future, the possibility of obtaining two additional significant contributions is envisioned. Firstly, there is a goal to enhance the network architecture itself. The current approach employs a fixed Convolutional Neural Network (CNN) topology, yet research has demonstrated that adopting a more flexible approach in network design holds the potential for improved results. Exploring and incorporating such flexibility into the network structure represents a valuable avenue for further investigation. Secondly, there is an aspiration to delve deeper into the underlying mechanisms of emerging EMO approaches, with a specific focus on the challenging domain of neurotrajectory analysis. NSGA-III serves as an example of one such approach. By thoroughly studying and comprehending the intricacies of these novel EMO techniques, an opportunity arises to contribute significantly to the field. By pursuing these research directions, the aim is to expand the scope and impact of this work, thereby paving the way for future advancements in network architecture and the understanding of EMO methodologies within the context of neurotrajectory analysis.

 Moving forward, our intention is to apply our approach to various neurotrajectory benchmark problems that currently lack standardisation and widespread adoption within the research community. This undertaking is crucial in order to further illuminate the intricacies of this challenging problem and see the implications of using a particular EMO approach. By extending the application of our approach to these non-standardised benchmarks, we aim to contribute to the ongoing efforts of addressing the complexities associated with neurotrajectory analysis. Through our research, we aspire to foster greater understanding and advancements in this domain, ultimately paving the way for future breakthroughs and innovations.

	\section*{Acknowledgments}
	
	
	\noindent This publication has emanated from research conducted with the financial support of Science Foundation Ireland under Grant number 18/CRT/6049. The opinions, findings and conclusions or recommendations expressed in this material are those of the author(s) and do not necessarily reflect the views of the Science Foundation Ireland. The authors wish to acknowledge the Irish Centre for High-End Computing (ICHEC) for the provision of computational facilities and support.

	\bibliography{biblio_erc,ref,neuroevolution}

\begin{thebibliography}{10}
\expandafter\ifx\csname url\endcsname\relax
  \def\url#1{\texttt{#1}}\fi
\expandafter\ifx\csname urlprefix\endcsname\relax\def\urlprefix{URL }\fi
\expandafter\ifx\csname href\endcsname\relax
  \def\href#1#2{#2} \def\path#1{#1}\fi

\bibitem{Goodfellow-et-al-2016}
I.~Goodfellow, Y.~Bengio, A.~Courville, Deep Learning, MIT Press, 2016.

\bibitem{DBLP:journals/nature/LeCunBH15}
Y.~LeCun, Y.~Bengio, G.~E. Hinton,
  \href{https://doi.org/10.1038/nature14539}{Deep learning}, Nature 521~(7553)
  (2015) 436--444.
\newblock \href {http://dx.doi.org/10.1038/nature14539}
  {\path{doi:10.1038/nature14539}}.
\newline\urlprefix\url{https://doi.org/10.1038/nature14539}

\bibitem{galvan2020neuroevolution}
E.~Galvan, P.~Mooney, Neuroevolution in deep neural networks: Current trends
  and future challenges, IEEE Transactions on Artificial Intelligence (2021)
  1--1\href {http://dx.doi.org/10.1109/TAI.2021.3067574}
  {\path{doi:10.1109/TAI.2021.3067574}}.

\bibitem{DBLP:conf/cvpr/SzegedyLJSRAEVR15}
C.~Szegedy, W.~Liu, Y.~Jia, P.~Sermanet, S.~E. Reed, D.~Anguelov, D.~Erhan,
  V.~Vanhoucke, A.~Rabinovich,
  \href{https://doi.org/10.1109/CVPR.2015.7298594}{Going deeper with
  convolutions}, in: {IEEE} Conference on Computer Vision and Pattern
  Recognition, {CVPR} 2015, Boston, MA, USA, June 7-12, 2015, {IEEE} Computer
  Society, 2015, pp. 1--9.
\newblock \href {http://dx.doi.org/10.1109/CVPR.2015.7298594}
  {\path{doi:10.1109/CVPR.2015.7298594}}.
\newline\urlprefix\url{https://doi.org/10.1109/CVPR.2015.7298594}

\bibitem{DBLP:conf/gecco/DengSG22}
S.~Deng, Y.~Sun, E.~Galv{\'{a}}n,
  \href{https://doi.org/10.1145/3520304.3528884}{Neural architecture search
  using genetic algorithm for facial expression recognition}, in: J.~E.
  Fieldsend, M.~Wagner (Eds.), {GECCO} '22: Genetic and Evolutionary
  Computation Conference, Companion Volume, Boston, Massachusetts, USA, July 9
  - 13, 2022, {ACM}, 2022, pp. 423--426.
\newblock \href {http://dx.doi.org/10.1145/3520304.3528884}
  {\path{doi:10.1145/3520304.3528884}}.
\newline\urlprefix\url{https://doi.org/10.1145/3520304.3528884}

\bibitem{Yanan2023}
S.~Deng, Z.~Lv, E.~Galv{\'{a}}n, Y.~Sun, Evolutionary neural architecture
  search for facial expression recognition, IEEE Transactions on Emerging
  Topics in Computational Intelligence.

\bibitem{Silver_2016}
D.~Silver, A.~Huang, C.~J. Maddison, A.~Guez, L.~Sifre, G.~van~den Driessche,
  J.~Schrittwieser, I.~Antonoglou, V.~Panneershelvam, M.~Lanctot, S.~Dieleman,
  D.~Grewe, J.~Nham, N.~Kalchbrenner, I.~Sutskever, T.~Lillicrap, M.~Leach,
  K.~Kavukcuoglu, T.~Graepel, D.~Hassabis, Mastering the game of {Go} with deep
  neural networks and tree search, Nature 529~(7587) (2016) 484--489.
\newblock \href {http://dx.doi.org/10.1038/nature16961}
  {\path{doi:10.1038/nature16961}}.

\bibitem{Back:1996:EAT:229867}
T.~B\"{a}ck, Evolutionary Algorithms in Theory and Practice: Evolution
  Strategies, Evolutionary Programming, Genetic Algorithms, Oxford University
  Press, Oxford, UK, 1996.

\bibitem{EibenBook2003}
A.~E. Eiben, J.~E. Smith, Introduction to {E}volutionary {C}omputing, Springer
  Verlag, 2003.

\bibitem{Eiben:2015:nature}
A.~E. Eiben, J.~Smith, From evolutionary computation to the evolution of
  things, Nature 521 (2015) 476--482.
\newblock \href {http://dx.doi.org/doi:10.1038/nature14544}
  {\path{doi:doi:10.1038/nature14544}}.

\bibitem{koza_2003}
J.~Koza, M.~Keane, M.~Streeter, W.~Mydlowec, J.~Yu, G.~Lanza, Genetic
  programming iv: Routine human-competitive machine intelligence.

\bibitem{Koza:2010:HRP:1831229.1831232}
J.~R. Koza,
  \href{http://dx.doi.org/10.1007/s10710-010-9112-3}{Human-competitive results
  produced by genetic programming}, Genetic Programming and Evolvable Machines
  11~(3-4) (2010) 251--284.
\newblock \href {http://dx.doi.org/10.1007/s10710-010-9112-3}
  {\path{doi:10.1007/s10710-010-9112-3}}.
\newline\urlprefix\url{http://dx.doi.org/10.1007/s10710-010-9112-3}

\bibitem{Bottou2012}
L.~Bottou, \href{https://doi.org/10.1007/978-3-642-35289-8_25}{Stochastic
  Gradient Descent Tricks}, Springer Berlin Heidelberg, Berlin, Heidelberg,
  2012, pp. 421--436.
\newblock \href {http://dx.doi.org/10.1007/978-3-642-35289-8_25}
  {\path{doi:10.1007/978-3-642-35289-8_25}}.
\newline\urlprefix\url{https://doi.org/10.1007/978-3-642-35289-8_25}

\bibitem{HECHTNIELSEN199265}
R.~Hecht-Nielsen,
  \href{https://www.sciencedirect.com/science/article/pii/B9780127412528500108}{Iii.3
  - theory of the backpropagation neural network**based on “nonindent” by
  robert hecht-nielsen, which appeared in proceedings of the international
  joint conference on neural networks 1, 593–611, june 1989. © 1989 ieee.},
  in: H.~Wechsler (Ed.), Neural Networks for Perception, Academic Press, 1992,
  pp. 65--93.
\newblock \href
  {http://dx.doi.org/https://doi.org/10.1016/B978-0-12-741252-8.50010-8}
  {\path{doi:https://doi.org/10.1016/B978-0-12-741252-8.50010-8}}.
\newline\urlprefix\url{https://www.sciencedirect.com/science/article/pii/B9780127412528500108}

\bibitem{726791}
Y.~{LeCun}, L.~{Bottou}, Y.~{Bengio}, P.~{Haffner}, Gradient-based learning
  applied to document recognition, Proceedings of the IEEE 86~(11) (1998)
  2278--2324.

\bibitem{10.1145/3065386}
A.~Krizhevsky, I.~Sutskever, G.~E. Hinton,
  \href{https://doi.org/10.1145/3065386}{Imagenet classification with deep
  convolutional neural networks}, Commun. ACM 60~(6) (2017) 84–90.
\newblock \href {http://dx.doi.org/10.1145/3065386}
  {\path{doi:10.1145/3065386}}.
\newline\urlprefix\url{https://doi.org/10.1145/3065386}

\bibitem{iet:/content/conferences/10.1049/cp_19991218}
F.~Gers,
  \href{https://digital-library.theiet.org/content/conferences/10.1049/cp_19991218}{Learning
  to forget: continual prediction with lstm}, IET Conference Proceedings (1999)
  850--855(5).
\newline\urlprefix\url{https://digital-library.theiet.org/content/conferences/10.1049/cp_19991218}

\bibitem{Deb:2001:MOU:559152}
K.~Deb, Multi-Objective Optimization Using Evolutionary Algorithms, John Wiley
  \& Sons, Inc., New York, NY, USA, 2001.

\bibitem{4358754}
Q.~Zhang, H.~Li, Moea/d: A multiobjective evolutionary algorithm based on
  decomposition, IEEE Transactions on Evolutionary Computation 11~(6) (2007)
  712--731.
\newblock \href {http://dx.doi.org/10.1109/TEVC.2007.892759}
  {\path{doi:10.1109/TEVC.2007.892759}}.

\bibitem{996017}
K.~{Deb}, A.~{Pratap}, S.~{Agarwal}, T.~{Meyarivan}, A fast and elitist
  multiobjective genetic algorithm: Nsga-ii, IEEE Transactions on Evolutionary
  Computation 6~(2) (2002) 182--197.
\newblock \href {http://dx.doi.org/10.1109/4235.996017}
  {\path{doi:10.1109/4235.996017}}.

\bibitem{8998284}
Q.~Xu, Z.~Xu, T.~Ma, A survey of multiobjective evolutionary algorithms based
  on decomposition: Variants, challenges and future directions, IEEE Access 8
  (2020) 41588--41614.
\newblock \href {http://dx.doi.org/10.1109/ACCESS.2020.2973670}
  {\path{doi:10.1109/ACCESS.2020.2973670}}.

\bibitem{Zitzler2001SPEA2IT}
E.~Zitzler, M.~Laumanns, L.~Thiele, Spea2: Improving the strength pareto
  evolutionary algorithm, 2001.

\bibitem{coello2004applications}
C.~A.~C. Coello, G.~B. Lamont, Applications of multi-objective evolutionary
  algorithms, Vol.~1, World Scientific, 2004.

\bibitem{GalvanSig2021}
E.~Galv{\'{a}}n, Neuroevolution in deep neural networks: A comprehensive
  survey, SIGEvolution 14~(1) (2021) 3--7.

\bibitem{1597059}
C.~A. Coello~Coello, Evolutionary multi-objective optimization: a historical
  view of the field, IEEE Computational Intelligence Magazine 1~(1) (2006)
  28--36.
\newblock \href {http://dx.doi.org/10.1109/MCI.2006.1597059}
  {\path{doi:10.1109/MCI.2006.1597059}}.

\bibitem{CoelloCoello1999}
C.~A. Coello~Coello, A comprehensive survey of evolutionary-based
  multiobjective optimization techniques, Knowledge and Information systems
  1~(3) (1999) 269--308.

\bibitem{Kim2017NEMON}
Y.-H. Kim, B.~Reddy, S.~Yun, C.~Seo, Nemo : Neuro-evolution with multiobjective
  optimization of deep neural network for speed and accuracy, 2017.

\bibitem{deng2012mnist}
L.~Deng, The mnist database of handwritten digit images for machine learning
  research, IEEE Signal Processing Magazine 29~(6) (2012) 141--142.

\bibitem{krizhevsky2009learning}
A.~Krizhevsky, G.~Hinton, et~al., Learning multiple layers of features from
  tiny images.

\bibitem{10.1145/3321707.3321729}
Z.~Lu, I.~Whalen, V.~Boddeti, Y.~Dhebar, K.~Deb, E.~Goodman, W.~Banzhaf,
  \href{https://doi.org/10.1145/3321707.3321729}{Nsga-net: Neural architecture
  search using multi-objective genetic algorithm}, in: Proceedings of the
  Genetic and Evolutionary Computation Conference, GECCO ’19, Association for
  Computing Machinery, New York, NY, USA, 2019, p. 419–427.
\newblock \href {http://dx.doi.org/10.1145/3321707.3321729}
  {\path{doi:10.1145/3321707.3321729}}.
\newline\urlprefix\url{https://doi.org/10.1145/3321707.3321729}

\bibitem{lu2020multiobjective}
Z.~Lu, I.~Whalen, Y.~Dhebar, K.~Deb, E.~D. Goodman, W.~Banzhaf, V.~N. Boddeti,
  Multiobjective evolutionary design of deep convolutional neural networks for
  image classification, IEEE Transactions on Evolutionary Computation 25~(2)
  (2020) 277--291.

\bibitem{xiao2017fashion}
H.~Xiao, K.~Rasul, R.~Vollgraf, Fashion-mnist: a novel image dataset for
  benchmarking machine learning algorithms, arXiv preprint arXiv:1708.07747.

\bibitem{netzer2011reading}
Y.~Netzer, T.~Wang, A.~Coates, A.~Bissacco, B.~Wu, A.~Y. Ng, Reading digits in
  natural images with unsupervised feature learning.

\bibitem{elsken2019openreview}
T.~Elsken, J.~H. Metzen, F.~Hutter,
  \href{https://openreview.net/forum?id=ByME42AqK7}{Efficient multi-objective
  neural architecture search via lamarckian evolution}, in: 7th International
  Conference on Learning Representations, {ICLR} 2019, New Orleans, LA, USA,
  May 6-9, 2019, OpenReview.net, 2019.
\newline\urlprefix\url{https://openreview.net/forum?id=ByME42AqK7}

\bibitem{loni2020deepmaker}
M.~Loni, S.~Sinaei, A.~Zoljodi, M.~Daneshtalab, M.~Sj{\"o}din, Deepmaker: A
  multi-objective optimization framework for deep neural networks in embedded
  systems, Microprocessors and Microsystems 73 (2020) 102989.

\bibitem{lu2023neural}
Z.~Lu, R.~Cheng, Y.~Jin, K.~C. Tan, K.~Deb, Neural architecture search as
  multiobjective optimization benchmarks: Problem formulation and performance
  assessment, IEEE Transactions on Evolutionary Computation.

\bibitem{he2021survey}
L.~He, H.~Ishibuchi, A.~Trivedi, H.~Wang, Y.~Nan, D.~Srinivasan, A survey of
  normalization methods in multiobjective evolutionary algorithms, IEEE
  Transactions on Evolutionary Computation 25~(6) (2021) 1028--1048.

\bibitem{miret2022neuroevolution}
S.~Miret, V.~S. Chua, M.~Marder, M.~Phiellip, N.~Jain, S.~Majumdar,
  Neuroevolution-enhanced multi-objective optimization for mixed-precision
  quantization, in: Proceedings of the Genetic and Evolutionary Computation
  Conference, 2022, pp. 1057--1065.

\bibitem{deb2013evolutionary}
K.~Deb, H.~Jain, An evolutionary many-objective optimization algorithm using
  reference-point-based nondominated sorting approach, part i: solving problems
  with box constraints, IEEE transactions on evolutionary computation 18~(4)
  (2013) 577--601.

\bibitem{liu2020multi}
Q.~Liu, X.~Li, H.~Liu, Z.~Guo, Multi-objective metaheuristics for discrete
  optimization problems: A review of the state-of-the-art, Applied Soft
  Computing 93 (2020) 106382.

\bibitem{grigorescu2019ieee}
S.~Grigorescu, B.~Trasnea, L.~Marina, A.~Vasilcoi, T.~Cocias, Neurotrajectory:
  A neuroevolutionary approach to local state trajectory learning for
  autonomous vehicles, IEEE Robotics and Automation Letters PP (2019) 1--1.
\newblock \href {http://dx.doi.org/10.1109/LRA.2019.2926224}
  {\path{doi:10.1109/LRA.2019.2926224}}.

\bibitem{fox1997dynamic}
D.~Fox, W.~Burgard, S.~Thrun, The dynamic window approach to collision
  avoidance, IEEE Robotics \& Automation Magazine 4~(1) (1997) 23--33.

\bibitem{bojarski2016end}
M.~Bojarski, D.~Del~Testa, D.~Dworakowski, B.~Firner, B.~Flepp, P.~Goyal, L.~D.
  Jackel, M.~Monfort, U.~Muller, J.~Zhang, et~al., End to end learning for
  self-driving cars, arXiv preprint arXiv:1604.07316.

\bibitem{stapleton2022neuroevolutionary}
F.~Stapleton, E.~Galv\'{a}n, G.~Sistu, S.~Yogamani,
  \href{https://doi.org/10.1145/3520304.3528984}{Neuroevolutionary
  multi-objective approaches to trajectory prediction in autonomous vehicles},
  in: Proceedings of the Genetic and Evolutionary Computation Conference
  Companion, GECCO '22, Association for Computing Machinery, New York, NY, USA,
  2022, p. 675–678.
\newblock \href {http://dx.doi.org/10.1145/3520304.3528984}
  {\path{doi:10.1145/3520304.3528984}}.
\newline\urlprefix\url{https://doi.org/10.1145/3520304.3528984}

\bibitem{doi:10.1113/jphysiol.1962.sp006837}
D.~H. Hubel, T.~N. Wiesel,
  \href{https://physoc.onlinelibrary.wiley.com/doi/abs/10.1113/jphysiol.1962.sp006837}{Receptive
  fields, binocular interaction and functional architecture in the cat's visual
  cortex}, The Journal of Physiology 160~(1) (1962) 106--154.
\newblock \href
  {http://arxiv.org/abs/https://physoc.onlinelibrary.wiley.com/doi/pdf/10.1113/jphysiol.1962.sp006837}
  {\path{arXiv:https://physoc.onlinelibrary.wiley.com/doi/pdf/10.1113/jphysiol.1962.sp006837}},
  \href {http://dx.doi.org/10.1113/jphysiol.1962.sp006837}
  {\path{doi:10.1113/jphysiol.1962.sp006837}}.
\newline\urlprefix\url{https://physoc.onlinelibrary.wiley.com/doi/abs/10.1113/jphysiol.1962.sp006837}

\bibitem{Lecun98gradient-basedlearning}
Y.~Lecun, L.~Bottou, Y.~Bengio, P.~Haffner, Gradient-based learning applied to
  document recognition, in: Proceedings of the IEEE, 1998, pp. 2278--2324.

\bibitem{dos-santos-gatti-2014-deep}
C.~dos Santos, M.~Gatti, \href{https://www.aclweb.org/anthology/C14-1008}{Deep
  convolutional neural networks for sentiment analysis of short texts}, in:
  Proceedings of {COLING} 2014, the 25th International Conference on
  Computational Linguistics: Technical Papers, Dublin City University and
  Association for Computational Linguistics, Dublin, Ireland, 2014, pp. 69--78.
\newline\urlprefix\url{https://www.aclweb.org/anthology/C14-1008}

\bibitem{10.1145/3219819.3219890}
R.~Ying, R.~He, K.~Chen, P.~Eksombatchai, W.~L. Hamilton, J.~Leskovec,
  \href{https://doi.org/10.1145/3219819.3219890}{Graph convolutional neural
  networks for web-scale recommender systems}, in: Proceedings of the 24th ACM
  SIGKDD International Conference on Knowledge Discovery \& Data Mining, KDD
  ’18, Association for Computing Machinery, New York, NY, USA, 2018, p.
  974–983.
\newblock \href {http://dx.doi.org/10.1145/3219819.3219890}
  {\path{doi:10.1145/3219819.3219890}}.
\newline\urlprefix\url{https://doi.org/10.1145/3219819.3219890}

\bibitem{7508408}
K.~{Greff}, R.~K. {Srivastava}, J.~{Koutník}, B.~R. {Steunebrink},
  J.~{Schmidhuber}, Lstm: A search space odyssey, IEEE Transactions on Neural
  Networks and Learning Systems 28~(10) (2017) 2222--2232.

\bibitem{963769}
F.~A. {Gers}, E.~{Schmidhuber}, Lstm recurrent networks learn simple
  context-free and context-sensitive languages, IEEE Transactions on Neural
  Networks 12~(6) (2001) 1333--1340.

\bibitem{minkowski1910geometrie}
H.~Minkowski, Geometrie der zahlen, BG Teubner, 1910.

\bibitem{DBLP:conf/gecco/GalvanS19}
E.~Galv{\'{a}}n, M.~Schoenauer,
  \href{https://doi.org/10.1145/3321707.3321854}{Promoting semantic diversity
  in multi-objective genetic programming}, in: A.~Auger, T.~St{\"{u}}tzle
  (Eds.), Proceedings of the Genetic and Evolutionary Computation Conference,
  {GECCO} 2019, Prague, Czech Republic, July 13-17, 2019, {ACM}, 2019, pp.
  1021--1029.
\newblock \href {http://dx.doi.org/10.1145/3321707.3321854}
  {\path{doi:10.1145/3321707.3321854}}.
\newline\urlprefix\url{https://doi.org/10.1145/3321707.3321854}

\bibitem{DBLP:journals/asc/GalvanTS22}
E.~Galv{\'{a}}n, L.~Trujillo, F.~Stapleton,
  \href{https://doi.org/10.1016/j.asoc.2021.108143}{Semantics in
  multi-objective genetic programming}, Appl. Soft Comput. 115 (2022) 108143.
\newblock \href {http://dx.doi.org/10.1016/j.asoc.2021.108143}
  {\path{doi:10.1016/j.asoc.2021.108143}}.
\newline\urlprefix\url{https://doi.org/10.1016/j.asoc.2021.108143}

\bibitem{DBLP:conf/ssci/GalvanS20}
E.~Galv{\'{a}}n, F.~Stapleton,
  \href{https://doi.org/10.1109/SSCI47803.2020.9308386}{Semantic-based distance
  approaches in multi-objective genetic programming}, in: 2020 {IEEE} Symposium
  Series on Computational Intelligence, {SSCI} 2020, Canberra, Australia,
  December 1-4, 2020, {IEEE}, 2020, pp. 149--156.
\newblock \href {http://dx.doi.org/10.1109/SSCI47803.2020.9308386}
  {\path{doi:10.1109/SSCI47803.2020.9308386}}.
\newline\urlprefix\url{https://doi.org/10.1109/SSCI47803.2020.9308386}

\bibitem{DBLP:conf/cec/StapletonG21}
F.~Stapleton, E.~Galv{\'{a}}n,
  \href{https://doi.org/10.1109/CEC45853.2021.9504860}{Semantic neighborhood
  ordering in multi-objective genetic programming based on decomposition}, in:
  {IEEE} Congress on Evolutionary Computation, {CEC} 2021, Krak{\'{o}}w,
  Poland, June 28 - July 1, 2021, {IEEE}, 2021, pp. 580--587.
\newblock \href {http://dx.doi.org/10.1109/CEC45853.2021.9504860}
  {\path{doi:10.1109/CEC45853.2021.9504860}}.
\newline\urlprefix\url{https://doi.org/10.1109/CEC45853.2021.9504860}

\bibitem{gridsimCode}
B.~Trasnea, A.~Vasilcoi, C.~Pozna, S.~Grigorescu, Gridsim,
  \url{https://github.com/RovisLab/GridSim} (2019).

\bibitem{trasnea2019gridsim}
B.~Trasnea, L.~A. Marina, A.~Vasilcoi, C.~R. Pozna, S.~M. Grigorescu, Gridsim:
  a vehicle kinematics engine for deep neuroevolutionary control in autonomous
  driving, in: 2019 Third IEEE International Conference on Robotic Computing
  (IRC), IEEE, 2019, pp. 443--444.

\bibitem{buhet2020plop}
T.~Buhet, E.~Wirbel, A.~Bursuc, X.~Perrotton, Plop: Probabilistic polynomial
  objects trajectory planning for autonomous driving, 4th Conference on Robot
  Learning, CoRL 2020, 16-18 November 2020, Virtual Event / Cambridge, MA,
  {USA} 155 (2020) 329--338.

\bibitem{mersch2021maneuver}
B.~Mersch, T.~H{\"o}llen, K.~Zhao, C.~Stachniss, R.~Roscher, Maneuver-based
  trajectory prediction for self-driving cars using spatio-temporal
  convolutional networks, in: 2021 IEEE/RSJ International Conference on
  Intelligent Robots and Systems (IROS), IEEE, 2021, pp. 4888--4895.

\bibitem{chandra2020forecasting}
R.~Chandra, T.~Guan, S.~Panuganti, T.~Mittal, U.~Bhattacharya, A.~Bera,
  D.~Manocha, Forecasting trajectory and behavior of road-agents using spectral
  clustering in graph-lstms, IEEE Robotics and Automation Letters 5~(3) (2020)
  4882--4890.

\bibitem{Mo2020InteractionAwareTP}
X.~Mo, Y.~Xing, C.~Lv, Interaction-aware trajectory prediction of connected
  vehicles using cnn-lstm networks, IECON 2020 The 46th Annual Conference of
  the IEEE Industrial Electronics Society (2020) 5057--5062.

\bibitem{caesar2020nuscenes}
H.~Caesar, V.~Bankiti, A.~H. Lang, S.~Vora, V.~E. Liong, Q.~Xu, A.~Krishnan,
  Y.~Pan, G.~Baldan, O.~Beijbom, nuscenes: A multimodal dataset for autonomous
  driving, in: Proceedings of the IEEE/CVF conference on computer vision and
  pattern recognition, 2020, pp. 11621--11631.

\bibitem{colyar2007us}
J.~Colyar, J.~Halkias, Us highway 101 dataset, Federal Highway Administration
  (FHWA), Tech. Rep. FHWA-HRT-07-030 (2007) 27--69.

\bibitem{krajewski2018highd}
R.~Krajewski, J.~Bock, L.~Kloeker, L.~Eckstein, The highd dataset: A drone
  dataset of naturalistic vehicle trajectories on german highways for
  validation of highly automated driving systems, in: 2018 21st International
  Conference on Intelligent Transportation Systems (ITSC), IEEE, 2018, pp.
  2118--2125.

\bibitem{huang2018apolloscape}
X.~Huang, X.~Cheng, Q.~Geng, B.~Cao, D.~Zhou, P.~Wang, Y.~Lin, R.~Yang, The
  apolloscape dataset for autonomous driving, in: Proceedings of the IEEE
  conference on computer vision and pattern recognition workshops, 2018, pp.
  954--960.

\bibitem{chang2019argoverse}
M.-F. Chang, J.~Lambert, P.~Sangkloy, J.~Singh, S.~Bak, A.~Hartnett, D.~Wang,
  P.~Carr, S.~Lucey, D.~Ramanan, et~al., Argoverse: 3d tracking and forecasting
  with rich maps, in: Proceedings of the IEEE/CVF conference on computer vision
  and pattern recognition, 2019, pp. 8748--8757.

\bibitem{kesten2019lyft}
R.~Kesten, M.~Usman, J.~Houston, T.~Pandya, K.~Nadhamuni, A.~Ferreira, M.~Yuan,
  B.~Low, A.~Jain, P.~Ondruska, et~al., Lyft level 5 av dataset 2019,
  urlhttps://level5. lyft. com/dataset 1 (2019) 3.

\bibitem{neurotrajCode}
S.~Grigorescu, B.~Trasnea, L.~Marina, A.~Vasilcoi, T.~Cocias, Neurotrajectory,
  \url{https://github.com/RovisLab/NeuroTrajectory} (2019).

\bibitem{Moshagen2021FindingHD}
T.~Moshagen, N.~A. Adde, A.~N. Rajgopal, Finding hidden-feature depending laws
  inside a data set and classifying it using neural network, ArXiv
  abs/2101.10427.

\bibitem{rmsprop}
T.~Tieleman, G.~Hinton, et~al., Lecture 6.5-rmsprop: Divide the gradient by a
  running average of its recent magnitude, COURSERA: Neural networks for
  machine learning 4~(2) (2012) 26--31.

\bibitem{adam}
D.~Kingma, J.~Ba, Adam: A method for stochastic optimization, International
  Conference on Learning Representations.

\bibitem{nadam}
T.~Dozat, Incorporating nesterov momentum into adam.

\bibitem{lecun2012efficient}
Y.~A. LeCun, L.~Bottou, G.~B. Orr, K.-R. M{\"u}ller, Efficient backprop, in:
  Neural networks: Tricks of the trade, Springer, 2012, pp. 9--48.

\bibitem{adagrad}
J.~Duchi, E.~Hazan, Y.~Singer, Adaptive subgradient methods for online learning
  and stochastic optimization, J. Mach. Learn. Res. 12~(null) (2011)
  2121–2159.

\bibitem{adadelta}
M.~D. Zeiler, Adadelta: an adaptive learning rate method, arXiv preprint
  arXiv:1212.5701.

\bibitem{Ruder2016AnOO}
S.~Ruder, An overview of gradient descent optimization algorithms, ArXiv
  abs/1609.04747.

\bibitem{higgins2004introduction}
J.~Higgins, \href{https://books.google.ie/books?id=vhmFQgAACAAJ}{An
  Introduction to Modern Nonparametric Statistics}, Duxbury advanced series,
  Brooks/Cole, 2004.
\newline\urlprefix\url{https://books.google.ie/books?id=vhmFQgAACAAJ}

\bibitem{deb2006searching}
K.~Deb, D.~Saxena, et~al., Searching for pareto-optimal solutions through
  dimensionality reduction for certain large-dimensional multi-objective
  optimization problems, in: Proceedings of the world congress on computational
  intelligence (WCCI-2006), 2006, pp. 3352--3360.

\bibitem{10.1007/978-3-642-19893-9_12}
H.~Ishibuchi, Y.~Hitotsuyanagi, H.~Ohyanagi, Y.~Nojima, Effects of the
  existence of highly correlated objectives on the behavior of moea/d, in:
  R.~H.~C. Takahashi, K.~Deb, E.~F. Wanner, S.~Greco (Eds.), Evolutionary
  Multi-Criterion Optimization, Springer Berlin Heidelberg, Berlin, Heidelberg,
  2011, pp. 166--181.

\bibitem{zheng2019towards}
J.~Zheng, Y.~Kou, Z.~Jing, Q.~Wu, Towards many-objective optimization:
  objective analysis, multi-objective optimization and decision-making, IEEE
  Access 7 (2019) 93742--93751.

\bibitem{scott2015multivariate}
D.~W. Scott, Multivariate density estimation: theory, practice, and
  visualization, John Wiley \& Sons, 2015.

\end{thebibliography}
	
\end{document}